%% file: main.tex
\documentclass[final,5p,times,twocolumn]{elsarticle}
\usepackage{microtype}
\usepackage{graphicx}
\usepackage{caption}
\usepackage{subcaption}
\usepackage{adjustbox}
\usepackage{longtable}
\usepackage{multirow}
\usepackage{algorithmicx}
\usepackage[ruled]{algorithm}
\usepackage{algpseudocode}
\usepackage[table,xcdraw]{xcolor}
\usepackage{multicol}
\usepackage{float}
\usepackage{amssymb}
\usepackage{bbm}
\usepackage{xcolor}
\usepackage{soul}
\usepackage{amsmath}
\usepackage{hyperref}
\usepackage{algorithm}
\usepackage{algorithmicx}
\usepackage{tikz}
\usetikzlibrary{positioning, arrows.meta}
\hypersetup{
    colorlinks=true,
    linkcolor=blue,
    filecolor=magenta,      
    urlcolor=cyan
    }

\usepackage{booktabs}

\begin{document}

\begin{frontmatter}

\title{A Quad-Step Approach to Uncertainty-Aware Deep Learning for Skin Cancer Classification}

\author[inst1]{Hamzeh~Asgharnezhad}
\author[inst1]{Pegah~Tabarisaadi}
\author[inst1]{Abbas~Khosravi}
\author[inst1]{Roohallah~Alizadehsani}
\author[inst2]{U.~Rajendra~Acharya}

\affiliation[inst1]{organization={Deakin University, Australia (e-mail: abbas.khosravi@deakin.edu.au)}}
\affiliation[inst2]{organization=School of Mathematics, Physics and Computing, University of Southern Queensland, Toowoomba, QLD, Australia}

\begin{abstract}
Accurate skin cancer diagnosis is vital for early treatment and improved patient outcomes. Deep learning (DL) models have shown promise in automating skin cancer classification, yet challenges remain due to data scarcity and limited uncertainty awareness. This study presents a comprehensive evaluation of DL-based skin lesion classification with transfer learning and uncertainty quantification (UQ) on the HAM10000 dataset. We benchmark several pre-trained feature extractors—including CLIP variants, ResNet50, DenseNet121, VGG16, and EfficientNet-V2-Large—combined with traditional classifiers such as SVM, XGBoost, and logistic regression. Multiple principal component analysis (PCA) settings (64, 128, 256, 512) are explored, with LAION CLIP ViT-H/14 and ViT-L/14 at PCA-256 achieving the strongest baseline results. In the UQ phase, Monte Carlo Dropout (MCD), Ensemble, and Ensemble Monte Carlo Dropout (EMCD) are applied and evaluated using uncertainty-aware metrics (UAcc, USen, USpe, UPre). Ensemble methods with PCA-256 provide the best balance between accuracy and reliability. Further improvements are obtained through feature fusion of top-performing extractors at PCA-256.
Finally, we propose a feature-fusion–based model trained with a predictive entropy (PE) loss function, which outperforms all prior configurations across both standard and uncertainty-aware evaluations, advancing trustworthy DL-based skin cancer diagnosis.
\end{abstract}

\renewcommand\thefootnote{} % Remove footnote numbering


\end{frontmatter}
%-----------------------------------------------------------------------------
\section{Introduction}
Skin cancer is one of the most prevalent types of cancer worldwide, with its incidence rising significantly over the past decades. This increase is potentially linked to lifestyle changes and environmental factors such as ozone layer depletion \cite{jerant2000early}. Skin cancer primarily results from abnormal growth of skin cells and is categorized into two main types: melanoma and non-melanoma. While non-melanoma is more common and has a lower mortality rate, melanoma is responsible for nearly 75\% of skin cancer-related deaths \cite{shamsi2023novel}. In 2018, approximately 300,000 new cases of melanoma and over one million cases of non-melanoma were reported globally. Given current trends, these numbers are expected to rise substantially in the next two decades, making skin cancer a growing public health concern \cite{tabarisaadi2022uncertainty}.

Early diagnosis of skin cancer is crucial for improving survival rates and treatment outcomes. For instance, in the United States, the five-year survival rate for localized melanoma is 98.4\%, while for metastatic melanoma, it drops drastically to 22.5\% . The low mortality rate of non-melanoma skin cancer, at 0.64 deaths per 100,000 cases, further highlights the importance of early detection \cite{skrede2020deep}. However, diagnosing skin cancer remains a challenge, even for experienced dermatologists, as it requires careful analysis of dermatologist spot images. Given the projected increase in cases, relying solely on manual examination is not sustainable. Therefore, advanced diagnostic tools, such as artificial intelligence-based methods, are becoming increasingly essential to support healthcare professionals, enhance diagnostic accuracy, and reduce the burden on the medical system.

DL algorithms have emerged as powerful tools for skin cancer detection, offering significant improvements in accuracy, efficiency, and scalability compared to traditional diagnostic methods \cite{naveed2024pca, ramkumar2024novel}. Convolutional Neural Networks (CNNs), particularly, have demonstrated exceptional performance in analyzing dermatologist spot images by automatically learning complex patterns and distinguishing between malignant and benign lesions. These models can process large-scale datasets and extract hierarchical features that might be imperceptible to human experts, thus improving diagnostic reliability. Studies have shown that DL-based systems can achieve dermatologist-level performance in skin cancer classification, reducing the risk of misdiagnosis and enabling early detection, which is critical for successful treatment \cite{tabarisaadi2022uncertainty, skrede2020deep}.

Deep neural networks (DNNs) can struggle with unseen or ambiguous medical cases, potentially leading to incorrect diagnoses that may put patients at risk \cite{begoli2019need}. This limitation arises from their black-box nature, as they produce predictions without conveying the confidence behind them \cite{abdar2021review}. In medical diagnosis, particularly for critical conditions like skin cancer, understanding uncertainty is essential, as even a small misjudgment can have serious consequences. UQ helps bridge this gap by allowing AI models to estimate and communicate their confidence levels, empowering healthcare professionals to assess the reliability of predictions. By incorporating UQ, medical AI systems can prioritize uncertain cases for further evaluation, minimize diagnostic errors, and enhance transparency, ultimately improving patient safety and trust in automated diagnosis.

DL models require large amounts of labeled data to achieve high accuracy. However, obtaining such datasets in the medical domain remains a significant challenge \cite{litjens2017survey, cheplygina2019not, greenspan2016guest}. Unlike general computer vision tasks where massive labeled datasets are readily available, medical imaging datasets are often limited in size. Labeling medical data is not only time-consuming but also expensive, as it requires expert annotation from professionals \cite{esteva2017dermatologist}. Moreover, DNNs consist of thousands of parameters that must be optimized, and training them from scratch requires substantial computational resources and time. In fields like dermatology, the scarcity of labeled medical images further complicates training, increasing the risk of overfitting. These challenges emphasize the need for techniques to efficiently leverage limited data while ensuring high diagnostic performance.

One effective approach to overcoming the challenges of training deep learning models on limited medical datasets is transfer learning \cite{shin2016deep, tajbakhsh2016convolutional}. Instead of training a model from scratch, transfer learning leverages pre-trained networks that have already learned rich and generalizable features from large-scale datasets \cite{yosinski2014transferable}. These networks, trained initially on diverse image collections, can be fine-tuned or adapted to specific tasks, such as skin cancer detection, where labeled data is often scarce. By reusing learned representations, transfer learning significantly reduces computational costs, accelerates training, and enhances model generalization, making deep learning a feasible and effective solution for medical image analysis.

Several well-established transfer learning models have been widely used for image analysis due to their strong feature extraction capabilities. ResNet50 \cite{he2016deep} is known for its deep residual connections that improve gradient flow and allow the training of very deep networks. DenseNet121 \cite{huang2017densely} enhances feature propagation by densely connecting layers, leading to efficient parameter utilization. VGG16 \cite{simonyan2015very}, despite being an older architecture, remains popular due to its simple yet effective design. EfficientNet-V2-L \cite{tan2021efficientnetv2} builds upon compound scaling to optimize model size and performance, making it highly efficient for medical imaging tasks. CLIP \cite{radford2021learning}, a more recent approach, introduces contrastive learning with text-image supervision, potentially enabling a broader and more generalizable feature representation. These models differ in architectural depth, parameter efficiency, and feature extraction methods, which influence their suitability for specific tasks.

In this study, we explore the effectiveness of vision transformers for automated skin cancer classification through transfer learning. Specifically, we employ pre-trained transformer-based models, including various CLIP architectures, to extract deep features from dermoscopic images. These are compared against established convolutional neural network (CNN) feature extractors such as ResNet50, DenseNet121, VGG16, and EfficientNet-V2-L. To provide a more comprehensive evaluation of feature representation, we apply principal component analysis (PCA) with multiple component sizes, enabling a systematic investigation of the impact of dimensionality reduction. Our goal is to assess the representational capability of vision transformers relative to CNNs and to identify the feature-classifier-PCA combinations that are most effective for distinguishing between different types of skin cancer.

To classify the extracted features, we evaluate a diverse set of machine learning classifiers, including Support Vector Machine, Extreme Gradient Boosting, K-Nearest Neighbors, Logistic Regression, Random Forest, Gradient Boosting, Decision Tree, and Naive Bayes. This broad comparison enables us to investigate the interaction between feature representations, dimensionality reduction, and classification paradigms.

Beyond classification accuracy, uncertainty estimation is a critical aspect of medical AI applications. To ensure the reliability of automated diagnosis, we integrate uncertainty quantification (UQ) using methods such as model ensembles \cite{lakshminarayanan2017simple}, Monte Carlo Dropout (MCD) \cite{gal2016dropout}, and Ensemble Monte Carlo Dropout (EMCD). These approaches allow us to capture predictive uncertainty and assess confidence in model outputs, which is particularly important in clinical decision-making where incorrect predictions may have serious consequences.

Finally, building on the strengths of the best-performing configurations, we employ feature fusion of top extractors and propose a novel framework trained with a predictive entropy (PE) loss function. This work not only compares a wide range of feature extraction, dimensionality reduction, classification, and UQ methods, but also proposes the most effective configuration based on the results. The originality of our contribution lies in combining multi-scale PCA, feature fusion, and uncertainty-aware training with predictive entropy into a unified framework, advancing both the accuracy and reliability of automated skin cancer diagnosis.
The main contributions of this paper are as follows:
\begin{itemize}
    \item A comprehensive comparative analysis of CLIP-based vision transformers and CNN-based models for feature extraction in automated skin cancer classification.  
    \item Evaluation of multiple principal component analysis (PCA) settings to investigate the impact of dimensionality reduction on classification performance.  
    \item Assessment of a diverse set of machine learning classifiers on features extracted from both vision transformers and CNNs, leading to the identification of the most effective feature–classifier–PCA configurations.  
    \item Integration of uncertainty quantification (UQ) into the classification pipeline to support more reliable diagnostic outputs, with a detailed comparison of Monte Carlo Dropout, Deep Ensembles, and Ensemble MCD.  
    \item Development of a novel framework that combines feature fusion of top-performing models with a predictive entropy (PE) loss function, resulting in improved accuracy and uncertainty awareness.  
\end{itemize}
The remainder of this paper is organized as follows. Section~\ref{Sec:background} provides an overview of the transfer learning models, PCA configurations, and machine learning classifiers employed in this study, along with a detailed explanation of the uncertainty quantification methods and evaluation metrics used to assess prediction confidence. Section~\ref{Sec:Exp} introduces the HAM10000 dataset and presents a comprehensive experimental analysis, including performance comparisons across different feature extractor–classifier–PCA combinations, uncertainty-aware methods, and the proposed feature-fusion framework. Finally, Section~\ref{Sec:Conclusion} summarizes the key findings and highlights the originality of our work in combining multi-scale PCA, feature fusion, and predictive entropy–based uncertainty modeling for more accurate and trustworthy skin cancer diagnosis.
%--------------------------------------------------------------
\section{Background}\label{Sec:background}
\subsection{CLIP}
CLIP (Contrastive Language-Image Pre-training) is a vision model developed by OpenAI that learns transferable visual representations from natural language supervision \cite{radford2021learning}. Unlike traditional computer vision models trained on predefined object categories with labeled datasets, CLIP leverages large-scale natural language supervision by associating images with textual descriptions. The model is trained using a contrastive objective, where it learns to match an image with its corresponding textual description from a large dataset of (image, text) pairs. This approach enables CLIP to understand and generalize to various visual concepts without the need for dataset-specific fine-tuning. As a result, CLIP can perform zero-shot learning, meaning it can classify images into new categories based on text descriptions alone, without requiring additional labeled training data.

One of the key distinctions of CLIP from other pretrained algorithms, such as those trained on ImageNet, is its reliance on internet-scale text-image pairs rather than manually labeled datasets. The training dataset for CLIP consists of 400 million (image, text) pairs collected from the internet, referred to as WebImageText (WIT). This dataset is significantly larger and more diverse than conventional computer vision training datasets like ImageNet, which typically contain only a fixed set of labeled classes. The contrastive training objective allows CLIP to develop a broad understanding of visual concepts, enabling it to perform well across various  tasks, including optical character recognition (OCR), action recognition, and fine-grained object classification. Moreover, CLIP exhibits enhanced robustness to natural distribution shifts, making it more reliable in real-world applications where image distributions may differ from those in standard benchmark datasets.

CLIP effectiveness is demonstrated by its ability to perform competitively across a wide range of computer vision tasks without requiring additional fine-tuning. It achieves results comparable to fully supervised models on ImageNet while also transferring effectively to over 30 different benchmarks, including those involving object classification, scene recognition, and image retrieval. Its zero-shot capabilities allow it to classify images based on textual descriptions alone, making it highly adaptable to new tasks without the need for manually labeled training data. Additionally, CLIP exhibits strong robustness to domain shifts, maintaining better performance than traditional models when tested on datasets with natural variations. This adaptability, combined with its large-scale training approach, highlights its potential as a versatile tool for general-purpose vision applications.
\subsection{EfficientNetV2}
EfficientNetV2 \cite{tan2021efficientnetv2} is an improved convolutional network designed for faster training and better parameter efficiency compared to its predecessor, EfficientNet. It leverages training-aware neural architecture search and progressive learning to optimize training speed while maintaining high accuracy. A key innovation is the introduction of Fused-MBConv layers, which replace depthwise convolutions in early network stages to enhance computational efficiency. Unlike the original EfficientNet, which scaled depth, width, and resolution uniformly, EfficientNetV2 applies non-uniform scaling to allocate resources more effectively. These improvements result in models that train up to 11× faster and use 6.8× fewer parameters, making them highly suitable for real-world applications with limited computational resources.

EfficientNetV2 has been extensively trained on large-scale datasets, including ImageNet ILSVRC2012 (1.28 million images, 1,000 classes) and ImageNet21k (13 million images, 21,841 classes). Pretraining on ImageNet21k provides a substantial boost in generalization, allowing EfficientNetV2 to achieve 87.3\% top-1 accuracy on ImageNet ILSVRC2012 while maintaining computational efficiency. Beyond large-scale datasets, EfficientNetV2 has been tested on multiple transfer learning benchmarks, such as CIFAR-10, CIFAR-100, Flowers, and Cars, where it consistently outperforms prior models. The model ability to generalize well across different datasets highlights its robustness and adaptability in various domains, from everyday object classification to fine-grained recognition tasks.

Compared to other state-of-the-art models, such as ResNeSt, NFNets, and Vision Transformers (ViTs), EfficientNetV2 strikes a superior balance between accuracy, parameter size, and training speed. While ViTs have demonstrated strong performance on large datasets, they require extensive computational resources and long training times. In contrast, EfficientNetV2 achieves similar or better accuracy while training 5×–11× faster. Compared to NFNets and other convolutional architectures, EfficientNetV2 reduces training bottlenecks by dynamically adjusting image size and regularization strength, ensuring both efficiency and accuracy. These features make EfficientNetV2 a powerful choice for deep learning applications that require high performance with optimized resource utilization.
%--------------------------------------------------------------------------------
\subsection{Predictive Uncertainty Quantification}
Uncertainty quantification (UQ) plays a critical role in the deployment of machine learning models, particularly in high-stakes and real-world scenarios. While traditional models provide point predictions, they cannot often express their confidence in those outputs. This becomes problematic when models encounter unfamiliar, noisy, or ambiguous data, which are—common occurrences in practical applications. Without a reliable measure of uncertainty, users may blindly trust predictions, potentially leading to harmful or costly decisions. UQ addresses this gap by offering insights into the model confidence, enabling more informed decision-making, risk management, and system robustness. It allows stakeholders to identify when a prediction may be unreliable, trigger human oversight when needed, and even improve downstream processes by incorporating confidence levels into decision pipelines. Ultimately, integrating UQ not only enhances model transparency and trustworthiness but also helps bridge the gap between theoretical performance and real-world reliability.

In uncertainty-aware classification, it is often useful to define a threshold that separates predictions into certain and uncertain categories based on their associated uncertainty scores. This threshold enables practitioners to interpret the model’s outputs not only in terms of class labels but also in terms of how confident the model is in its decisions. Once this division is made, each prediction can be categorized into one of four groups: Correct and Certain (CC), Incorrect and Uncertain (IU), Correct and Uncertain (CU), and Incorrect and Certain (IC) \cite{tabarisaadi2022optimized}.

Ideally, a well-calibrated model should produce as many predictions as possible in the CC group, where the model is both correct and confident. The next desirable outcome is IU, where the model makes an error but acknowledges its uncertainty. This behavior is useful because it alerts the user to potential issues in the prediction, thereby preventing over-reliance on incorrect outputs. On the other hand, predictions in the CU category are less problematic but still suboptimal, as the model correctly classifies the input but lacks confidence in its own decision. The most undesirable group is IC, where the model makes a wrong prediction with high confidence—these cases are the most dangerous in real-world applications, as they may mislead users into making incorrect decisions with undue trust in the model.
To quantitatively evaluate the quality of uncertainty estimation, several metrics have been proposed based on these four groups. 
%-----------------------------------------------------------------------
\subsection{Monte Carlo Dropout (MCD)}
Monte Carlo Dropout is a widely adopted method for approximating Bayesian inference in neural networks. It addresses the challenge of computing the posterior distribution over model weights by using dropout not only during training but also at inference time, as proposed by Gal and Ghahramani~\cite{gal2016dropout}. This approach enables the network to produce multiple outputs by performing stochastic forward passes with randomly dropped neurons, effectively generating samples from an approximate posterior.

Given an input $x$, the mean prediction across $T$ such stochastic passes is given by:

\begin{equation}
\mu_{\text{pred}} \approx \frac{1}{T} \sum_{t=1}^{T} p(y = c \mid x, \hat{\omega}_t)
\end{equation}

\noindent where $p(y = c \mid x, \hat{\omega}_t)$ denotes the softmax probability for class $c$ obtained from the $t$-th forward pass using the sampled weights $\hat{\omega}_t$.

This method also supports the quantification of model uncertainty. A common measure is \textit{predictive entropy}, which evaluates the spread of the averaged class probabilities~\cite{gal2016dropout}:

\begin{equation} \label{Eq:MC-Dropout-PE}
PE = - \sum_c \mu_{\text{pred}} \log \mu_{\text{pred}}
\end{equation}

\noindent where the sum is over all possible output classes $c$. Lower entropy indicates higher confidence in the prediction, while higher entropy suggests greater uncertainty. This makes the metric particularly useful for identifying ambiguous or potentially unreliable predictions in classification tasks.
%----------------------------------------------------------------------
\begin{figure}[t]
\centering
\begin{subfigure}{0.6\columnwidth}
    \includegraphics[width=\columnwidth]{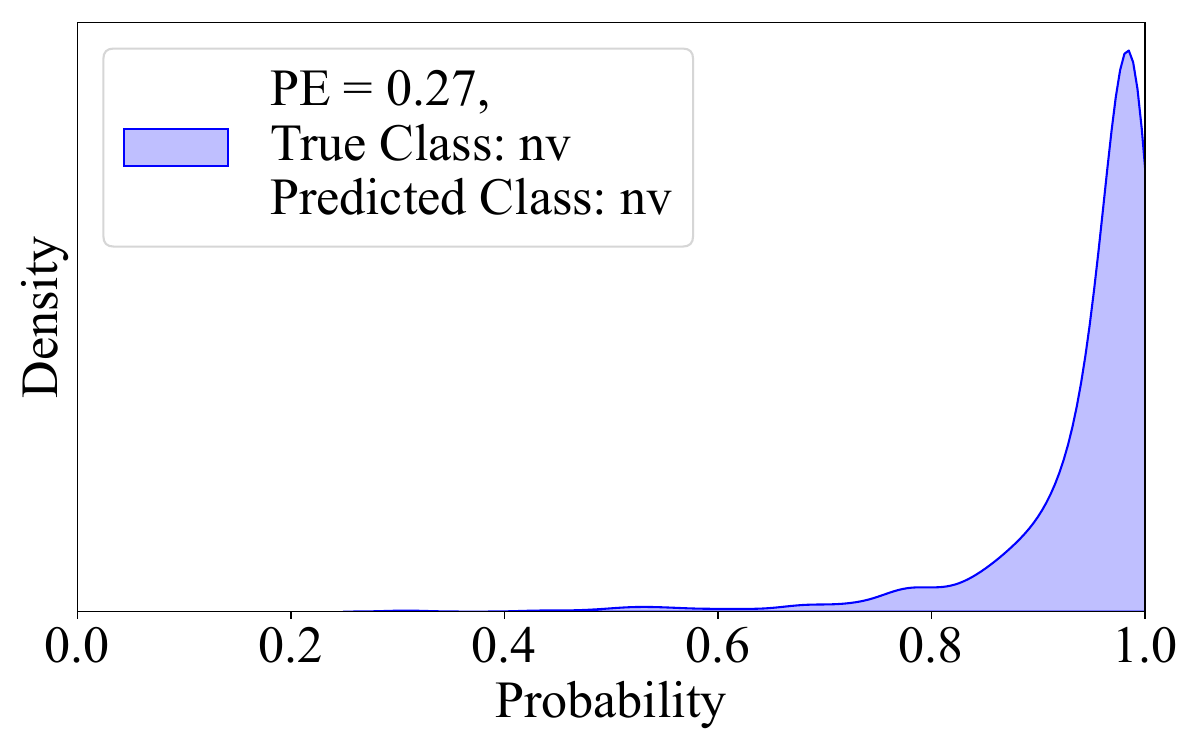}
    \caption{Certain Prediction}
    \label{fig:PE_Certain}
  \end{subfigure}
  \begin{subfigure}{0.6\columnwidth}
    \includegraphics[width=\columnwidth]{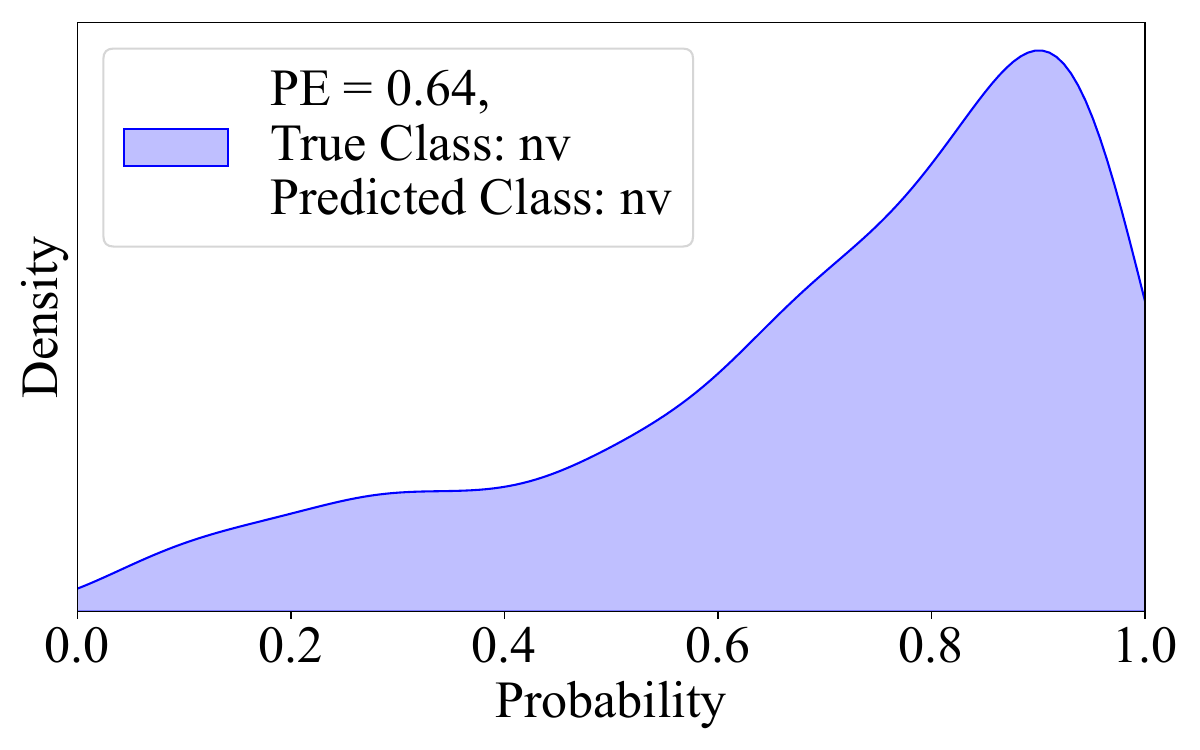}
    \caption{Uncertain Prediction} 
    \label{fig:PE_Uncertain}
  \end{subfigure}
  \caption{Comparison of Probability Density: Certain vs. Uncertain for test sample of class Nevus (LAION CLIP ViT-H/14 + MCD)}
  \label{fig:PE_Certain_Uncertain}
\end{figure}
%----------------------------------------------------------------------

Figure~\ref{fig:PE_Certain_Uncertain} illustrates the probability density distributions for two test samples classified as belonging to the Nevus class using the LAION CLIP ViT-H/14 model with Monte Carlo Dropout (MCD). Subfigure~\ref{fig:PE_Certain} depicts a confident prediction, where the distribution is sharply peaked around the predicted class, indicating low uncertainty. In contrast, Subfigure~\ref{fig:PE_Uncertain} shows a broader, flatter distribution, reflecting greater epistemic uncertainty in the model prediction.

%------------------------------------------------------------------------
\subsection{Deep Ensembles}
Deep ensembles provide a practical and effective means for estimating uncertainty by leveraging the diversity among multiple independently trained neural networks. Instead of relying on Bayesian inference, this method trains $N$ separate models using different random initializations and, optionally, varied data sampling strategies such as shuffling or bootstrapping~\cite{lakshminarayanan2017simple}. The variation in their learned parameters allows the ensemble to capture epistemic uncertainty through differences in model predictions.

For a given input $x$, each ensemble member generates a predictive distribution $p_{\theta_i}(y|x)$, where $\theta_i$ denotes the parameters of the $i^{\text{th}}$ model. The aggregated predictive distribution across the ensemble is calculated as:

\begin{equation}
\hat{p}(y|x) = \frac{1}{N} \sum_{i=1}^{N} p_{\theta_i}(y|x)
\end{equation}

To assess the uncertainty in the prediction, predictive entropy (PE) can be computed as:

\begin{equation}
PE = -\sum_{i=1}^{C} \hat{p}(y_i|x) \log \hat{p}(y_i|x)
\end{equation}

\noindent where $C$ is the number of classes in the classification task. A low entropy score corresponds to confident predictions, whereas a high entropy value signals uncertainty. As shown in the work by Lakshminarayanan et al.~\cite{lakshminarayanan2017simple}, this technique yields reliable and well-calibrated predictions without the need for complex Bayesian methods.
%------------------------------------------------------------------------
\subsection{Ensemble Monte Carlo Dropout (EMCD)}
Ensemble Monte Carlo Dropout (EMCD) is a hybrid strategy that merges the strengths of deep ensembles and Monte Carlo Dropout to enhance uncertainty estimation. In this method, $N$ separate neural networks are trained independently, as done in traditional ensembles. At inference time, each model generates $T$ stochastic predictions by applying dropout, similar to the Monte Carlo Dropout procedure.

For a test sample $x$, the $i^{\text{th}}$ model produces its predictive distribution by averaging the outputs from $T$ dropout-enabled forward passes:

\begin{equation}
\hat{p}_{\theta_i}(y|x) = \frac{1}{T} \sum_{t=1}^{T} p_{\theta_i^t}(y|x)
\end{equation}

\noindent where $p_{\theta_i^t}(y|x)$ refers to the softmax output obtained during the $t^{\text{th}}$ stochastic pass using a distinct dropout mask within the $i^{\text{th}}$ model.

To obtain the final predictive distribution from the EMCD model, the averaged outputs of all $N$ models are combined:

\begin{equation}
\hat{p}(y|x) = \frac{1}{N} \sum_{i=1}^{N} \hat{p}_{\theta_i}(y|x)
\end{equation}

The associated uncertainty can be quantified using predictive entropy:

\begin{equation}
PE = -\sum_{i=1}^{C} \hat{p}(y_i|x) \log \hat{p}(y_i|x)
\end{equation}

\noindent where $C$ represents the number of possible output classes. By integrating the model diversity introduced by ensembles with the randomness of dropout-based sampling, EMCD captures a richer representation of epistemic uncertainty and often leads to more calibrated predictions.
%----------------------------------------------------

\subsection{Uncertainty Confusion Matrix}
The uncertainty confusion matrix \cite{tabarisaadi2022optimized}, inspired by the traditional confusion matrix, is presented in Table~\ref{Tab:Unc-CF}. This matrix categorizes predictions based on both correctness (correct or incorrect) and the model’s confidence level (certain or uncertain). Ideally, an uncertainty-aware model should concentrate its predictions in the Correct and Certain (CC) and Incorrect and Uncertain (IU) categories, which are highlighted in green. These outcomes indicate confident correct predictions and cautious behavior when uncertain, respectively—both are desirable for safe and reliable decision-making. In contrast, predictions falling into the Incorrect and Certain (IC) and Correct and Uncertain (CU) categories (highlighted in red) are less favorable, particularly IC, which represents confident errors that may lead to misleading conclusions in real-world applications.

\begin{table}[t]
    \caption{Uncertainty confusion matrix}\label{Tab:Unc-CF}
    \centering
    \begin{tabular}{cc|c|c|}
    \cline{3-4}
                                                   &           & \multicolumn{2}{c|}{Correctness}                        \\ \cline{3-4} 
                                                   &           & Correct                    & Incorrect                  \\ \hline
\multicolumn{1}{|c|}{}                             & Certain   & \cellcolor[HTML]{9AFF99}CC & \cellcolor[HTML]{FD6864}IC \\ \cline{2-4} 
\multicolumn{1}{|c|}{\multirow{-2}{*}{Confidence}} & Uncertain & \cellcolor[HTML]{FD6864}CU & \cellcolor[HTML]{9AFF99}IU \\ \hline
    \end{tabular}
\end{table}

%------------------------------------------------------------------------
\subsubsection*{Uncertainty Sensitivity}
\textit{Uncertainty Sensitivity} (USen) is a key metric used to evaluate the effectiveness of a model uncertainty estimation. Specifically, it quantifies the model ability to recognize its own incorrect predictions as uncertain. In uncertainty-aware classification, this is highly desirable, as it ensures that erroneous predictions are accompanied by appropriate caution, thereby reducing the risk of overconfident failures in deployment.

The metric is defined as the proportion of incorrect predictions that the model also labels as uncertain. A higher value of $U_{sen}$ indicates better sensitivity to uncertainty in erroneous cases.

\begin{equation}
USen = \frac{N_{\text{IU}}}{N_{\text{IC}} + N_{\text{IU}}}
\end{equation}

\noindent where $N_{\text{IU}}$ is the number of predictions that are both Incorrect and Uncertain, and $N_{\text{incorrect}}$ is the total number of Incorrect predictions made by the model.
%----------------------------------------------------------------------------
\subsubsection*{Uncertainty Specificity}

\textit{Uncertainty Specificity} (USpe) evaluates the model’s ability to assign low uncertainty to its correct predictions. This metric reflects how confidently the model identifies its accurate outputs. In an ideal scenario, correct predictions should also be accompanied by high certainty, ensuring that the model confidence aligns with its correctness.

USpe is calculated as the proportion of predictions that are both Correct and Certain out of all Correct predictions made by the model.

\begin{equation}
USpe = \frac{N_{\text{CC}}}{N_{\text{CC}} + N_{\text{CU}}}
\end{equation}

\noindent where $N_{\text{CC}}$ denotes the number of Correct and Certain predictions, and $N_{\text{CU}}$ is the number of Correct and Uncertain predictions.
%---------------------------------------------------------------------------
\subsubsection*{Uncertainty Precision}

\textit{Uncertainty Precision} (UPre) measures how precise the model is in associating uncertainty with incorrect predictions. It indicates the proportion of uncertain predictions that are actually incorrect, helping to assess whether the model’s uncertainty estimates are meaningfully targeted at potentially erroneous outputs.

UPre is defined as the ratio of Incorrect and Uncertain predictions to the total number of Uncertain predictions:

\begin{equation}
UPre = \frac{N_{\text{IU}}}{N_{\text{CU}} + N_{\text{IU}}}
\end{equation}

\noindent where $N_{\text{IU}}$ is the number of Incorrect and Uncertain predictions, and $N_{\text{CU}}$ is the number of Correct and Uncertain predictions.
%---------------------------------------------------------
\subsubsection*{Uncertainty Accuracy}

\textit{Uncertainty Accuracy} (UAcc) provides an overall measure of how well the model aligns its uncertainty estimates with prediction correctness. Specifically, it captures the proportion of predictions that fall into the desired categories: either Correct and Certain (CC) or Incorrect and Uncertain (IU). These are considered ideal outcomes, where the model is confident when correct and cautious when wrong.

UAcc is calculated as the ratio of these desirable predictions to the total number of predictions made:

\begin{equation}
UAcc = \frac{N_{\text{CC}} + N_{\text{IU}}}{N_{\text{CC}}+N_{\text{CU}}+N_{\text{IU}}+N_{\text{IC}}}
\end{equation}

\noindent where $N_{\text{CC}}$ denotes the number of Correct and Certain predictions, $N_{\text{IU}}$ is the number of Incorrect and Uncertain predictions, and $N_{\text{CU}}$ is the number of correct and uncertain predictions and $N_{\text{IU}}$ is the number of incorrect and uncertain predictions made by the model.
%--------------------------------------------------
\subsection{Predictive Entropy Loss Function}
The predictive entropy, originally introduced as an uncertainty measure in Bayesian neural networks \cite{gal2016dropout}, can be incorporated into the training process as a regularization term. Recent studies have shown that adding predictive entropy to the training objective encourages the model to account for uncertainty during optimization \cite{shamsi2023novel, asgharnezhad2023improving}. Specifically, for each sample, the predictive entropy ($\mathrm{PE}_i$), estimated over $T$ stochastic forward passes, is added to the standard cross-entropy loss. For a batch of size $B$, the overall training objective is formulated as:

\begin{equation}
Loss_{batch} = CrossEntropy_{batch} + \frac{1}{B} \sum_{i=1}^{B} \mathrm{PE}_i
\end{equation}

This formulation combines the discriminative strength of cross-entropy with an additional penalty that reflects uncertainty. The entropy term encourages confident predictions when sufficient evidence exists, while preserving higher entropy for ambiguous inputs. As a result, the model achieves improved robustness, particularly when encountering test data that differs from the training distribution.

%--------------------------------------------------
\section{Simulation Results and Discussions}\label{Sec:Exp}
In this section, the proposed algorithm is applied to real-world datasets, and the performances of different algorithms are compared from different perspectives. 
%----------------------------------------------------------------------------
\subsection{Dataset}
The HAM10000 dataset \cite{tschandl2018ham10000}, short for 'Human Against Machine with 10000 training images', is a widely recognized resource for research in dermatological image analysis, especially in the development of machine learning models for skin lesion classification. It was compiled by researchers at the Medical University of Vienna and the Australian Skin Cancer Research Institute to provide a comprehensive and diverse collection of dermatoscopic images that reflect real-world clinical conditions. The dataset contains 10,015 dermatoscopic images representing seven types of pigmented skin lesions. These include melanocytic nevi (nv), melanoma (mel), benign keratosis-like lesions (bkl), basal cell carcinoma (bcc), actinic keratoses and intraepithelial carcinoma (akiec), vascular lesions (vasc), and dermatofibroma (df). Alongside the images, metadata such as the patient's age, sex, and lesion location are also provided, enabling the development of more nuanced and context-aware diagnostic models.
One of the strengths of HAM10000 is the variability in its sources. The images were gathered from two different geographic and clinical settings—Austria and Australia—using a range of dermatoscopic devices. This diversity makes the dataset particularly useful for training models that need to generalize well across different patient populations and imaging environments.

HAM10000 has become a foundational dataset in automated skin cancer detection and was prominently featured in the ISIC (International Skin Imaging Collaboration) 2018 challenge. It continues to support research in deep learning, uncertainty quantification, and explainable AI by providing a large, high-quality dataset that captures the complexity of dermatological diagnosis.

\begin{figure}[t]
  \centering
  \includegraphics[width=0.9\linewidth]{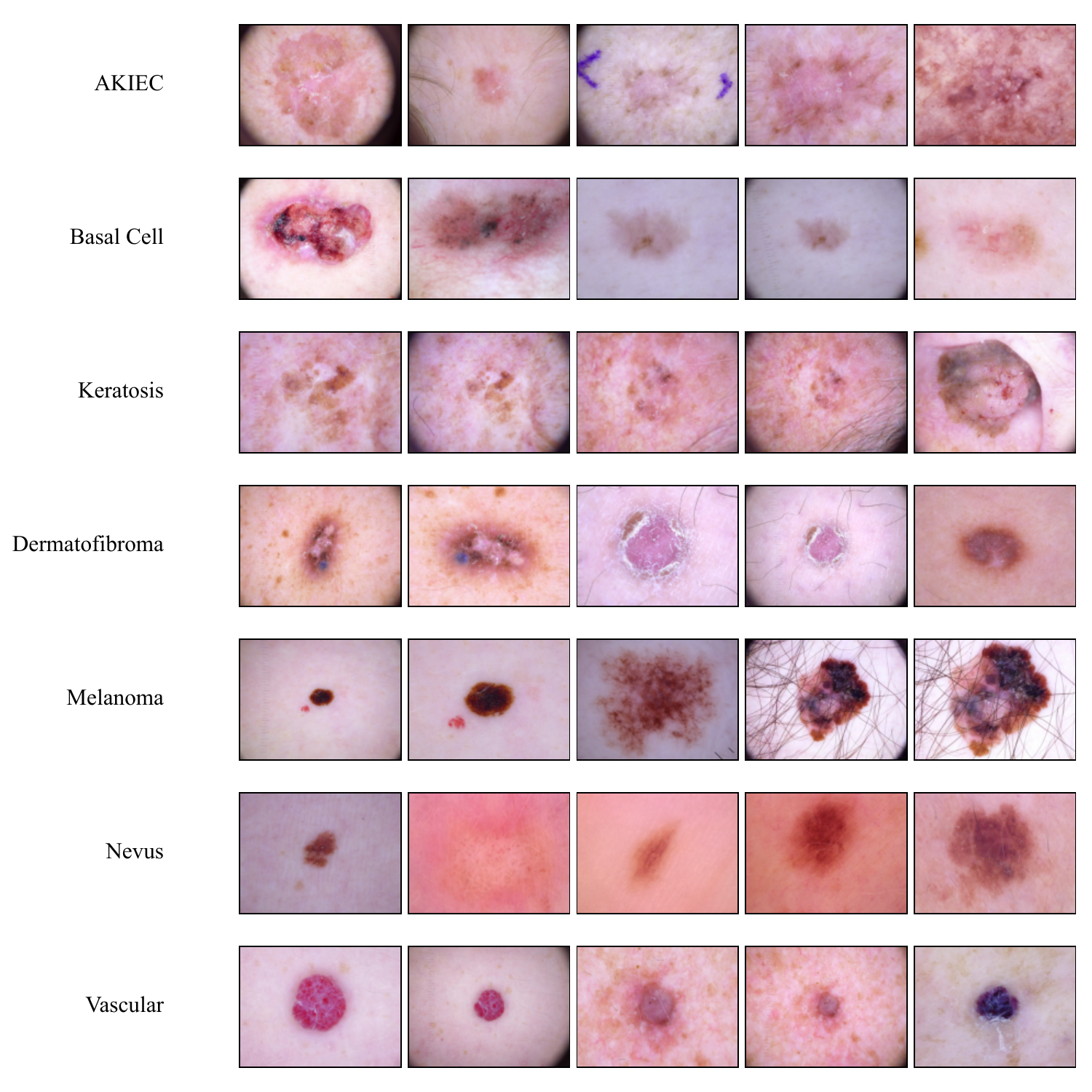}
  \caption{Examples of skin lesions from HAM10000 Dataset.}
  \label{fig:HAM10000}
\end{figure}

Figure~\ref{fig:HAM10000} illustrates the distribution of lesion types across the seven diagnostic categories included in the HAM10000 dataset. These classes are: Nevus, Melanoma, Keratosis, Basal Cell, AKIEC, Vascular and Dermatofibroma.
%----------------------------------------------------------------------------
%-------------------------------------------------
\subsection{Results} \label{sec:results}

%----------------------------------------------
\begin{figure*}[htbp]
  \centering
  \includegraphics[width=0.8\linewidth]{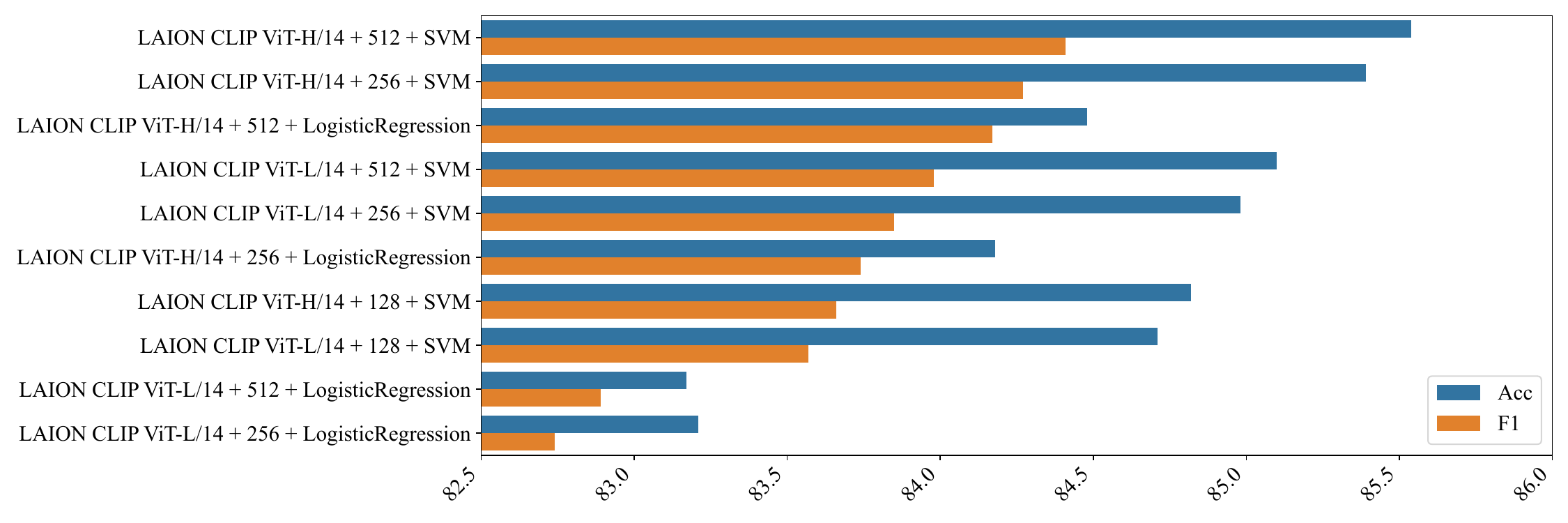}
    \caption{Performance comparison of the best-performing combinations of feature extractors and classifiers. Evaluation done based on Accuracy (Acc) and F1 Score (F1). The results reflect the impact of CLIP and ResNet-based extractors paired with classifiers such as SVM, XGBoost, k-Nearest Neighbors, and Logistic Regression.}
  \label{fig:performance_metrics_by_feature_extractor}
\end{figure*}
%----------------------------------------------------
To establish a strong baseline for skin lesion classification using the HAM10000 dataset, we conducted a comprehensive experimental comparison of state-of-the-art transfer learning feature extractors in combination with various traditional machine learning classifiers. The objective was to evaluate the performance of different visual representation models, including CLIP-based vision transformers, convolutional neural networks, and EfficientNet variants, when employed as fixed feature extractors. The extracted features were subsequently reduced to multiple dimensionalities—specifically 64, 100, 128, 256, and 512—using Principal Component Analysis (PCA)~\cite{jolliffe2016principal}. These reduced feature sets were then passed to a range of classifiers, including support vector machines (SVM), logistic regression, k-nearest neighbors, decision trees, random forests, XGBoost, gradient boosting, and naive Bayes. Each model configuration was evaluated using key metrics such as accuracy, precision, recall, and F1 Score (F1).

Table~\ref{tab:performance_metrics_by_feature_extractor} reports the mean results from 10 independent runs of each algorithm, providing robustness and reducing the influence of variability during training. Figure~\ref{fig:performance_metrics_by_feature_extractor} further illustrates the performance of the best-performing feature extractor–classifier combinations across the key metrics of accuracy and F1 Score. This visualization enables a clear comparative analysis of the most effective setups for skin lesion classification. The results reveal consistent trends across different configurations. In particular, the combination of the LAION CLIP ViT-H/14 feature extractor with 512-dimensional PCA features and an SVM classifier achieved the strongest overall performance, yielding an average accuracy of 85.54\%, precision of 84.81\%, recall of 85.54\%, and F1 Score of 84.41. These findings underscore the effectiveness of CLIP-based vision transformers—especially those pre-trained on large, diverse datasets such as LAION—in producing highly discriminative representations for medical image classification tasks like skin lesion detection.

In aggregate, SVM delivered the strongest and most consistent performance, especially with CLIP features at higher PCA dimensions (256–512). Logistic regression was a close second across many CLIP settings, while tree-based methods (notably XGBoost, and to a lesser extent Random Forests) trailed the linear models but remained competitive in the mid-to-upper 70s/low-80s for CLIP features. In contrast, k-nearest neighbors, decision trees, and naive Bayes generally underperformed—this gap was most pronounced with EfficientNet-V2-L features, where naive Bayes collapsed (e.g., 29.46–40.00\% accuracy at PCA 256–64) despite high class-precision artifacts. Overall, the results suggest that the high-dimensional, discriminative CLIP embeddings are best leveraged by margin-based or linear classifiers (SVM, logistic), whereas simpler or high-bias models struggle to model the complex decision boundaries.

Interestingly, although EfficientNet-V2-L is widely recognized for strong performance in conventional image classification benchmarks, it consistently underperformed in this setting, particularly when paired with simpler classifiers. This underscores the importance of not only selecting a capable feature extractor but also ensuring its compatibility with the downstream classifier. Overall, our findings demonstrate that CLIP-based vision transformers—most notably the ViT-H/14 variant—provide a highly effective foundation for skin lesion classification, achieving their best results when combined with margin-based classifiers such as SVM. 

In the second phase of our study, we extended our baseline experiments by incorporating uncertainty quantification into the classification pipeline. Building on the previously evaluated feature extractors, we applied three widely-used uncertainty-aware techniques: Monte Carlo Dropout (MCD), Ensemble, and Ensemble Monte Carlo Dropout (EMCD). We employed a custom neural network architecture, consisting of two hidden layers with 64 and 16 neurons respectively and ReLU activations, intermediate dropout layers to enable stochastic behavior during inference, and an output layer for classification. For the Ensemble and EMCD methods, six instances of this model were trained independently using different random weight initializations to capture model uncertainty. These methods were combined with various deep feature extractors to analyze not only classification performance but also the model’s behavior under uncertainty. The goal was to assess both predictive accuracy and the model’s ability to express and handle uncertainty in its outputs, which is particularly important for high-stakes applications like medical diagnosis.

Each uncertainty-aware method was applied to the extracted features without retraining the feature extractor. The classifier in each case was adapted to support uncertainty estimation. In addition to the standard evaluation metrics such as accuracy, precision, recall, and F1, we included several uncertainty-specific metrics: the number of correct and certain predictions (CC), Incorrect and certain predictions (IC), correct and uncertain predictions (CU), and Incorrect and uncertain predictions (IU), as well as uncertainty accuracy,  uncertainty sensitivity, uncertainty specificity, and uncertainty precision. These metrics help us understand not just how accurate a model is, but also how well it knows when it is uncertain.

\begin{figure*}[t]
  \centering
  \includegraphics[width=0.8\linewidth]{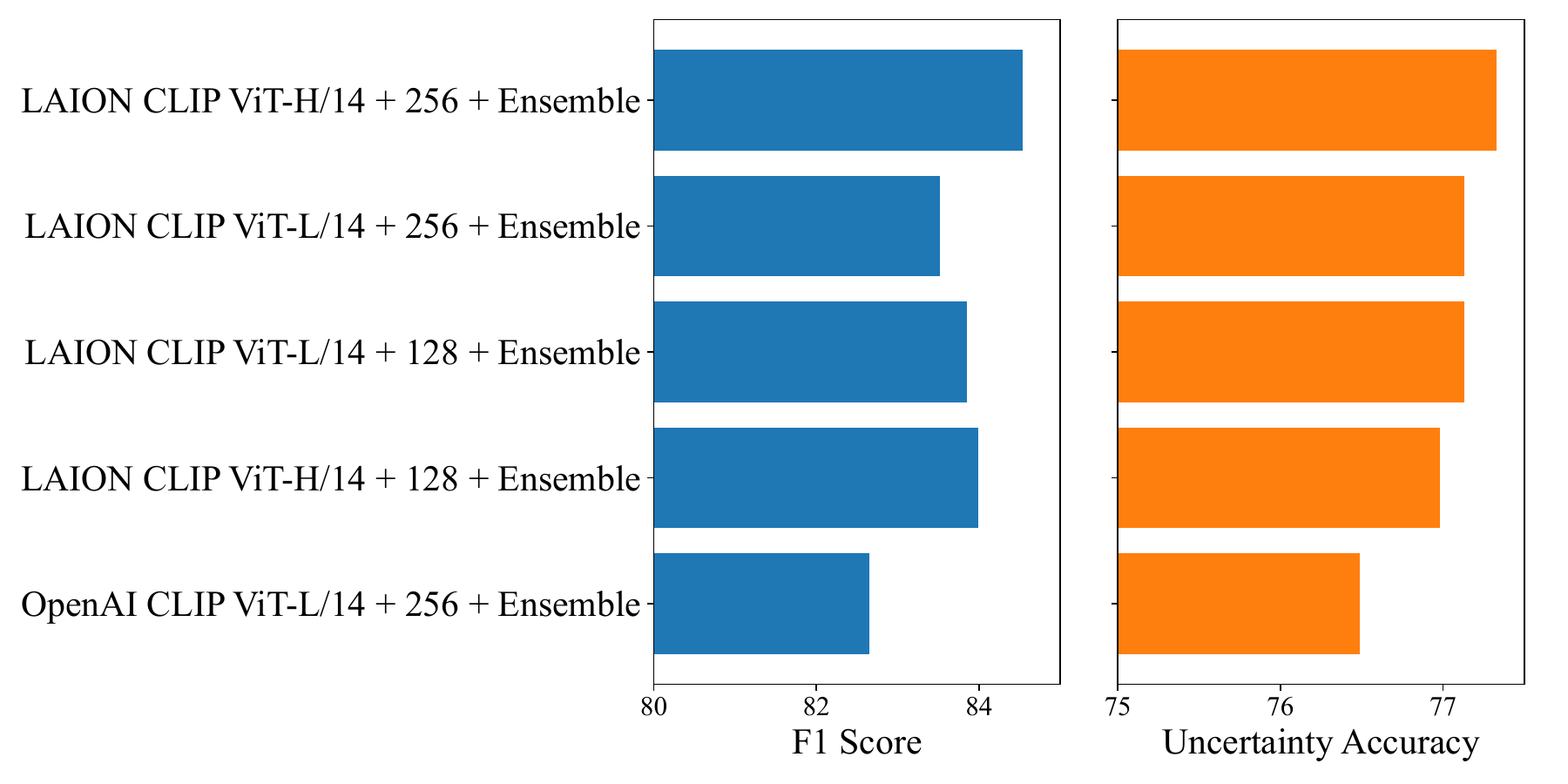}
  \caption{Performance comparison of the best-performing algorithms across various CLIP-based architectures and uncertainty-aware strategies. Evaluated metrics include F1 Score (F1) and Uncertainty Accuracy (UAcc).}
  \label{fig:uacc_f1}
\end{figure*}

Table~\ref{tab:uncertainty_results} show the results, while Figures~\ref{fig:uacc_f1} visualize the leading uncertainty-aware models. The highest performance was consistently achieved by Ensemble methods applied to CLIP-based vision transformers with PCA-256. In particular, the LAION CLIP ViT-H/14 + Ensemble model obtained the best overall uncertainty accuracy (UAcc = 77.33\%) while maintaining strong classification metrics (Acc = 85.47\%, F1 = 84.54\%) and producing the largest number of correct-and-certain predictions (CC = 1318). Very similar performance was observed with the LAION CLIP ViT-L/14 + Ensemble, which reached a UAcc of 77.13\%, an F1 of 83.52\%, and CC = 1299. Ensemble models with PCA-128 also performed competitively, such as the ViT-H/14 configuration with a UAcc of 76.98\%, combining balanced classification accuracy and relatively low rates of incorrect certainty. Taken together, these results highlight that CLIP-based feature extractors—particularly ViT-H/14 and ViT-L/14—paired with Ensemble uncertainty estimation at PCA dimensions of 128–256 provide the most effective balance between classification accuracy and reliable uncertainty quantification for skin lesion classification.
%-------------------------------------------------
%------------------------------------------------

\begin{figure*}[htbp]
  \centering
  \includegraphics[width=0.8\linewidth]{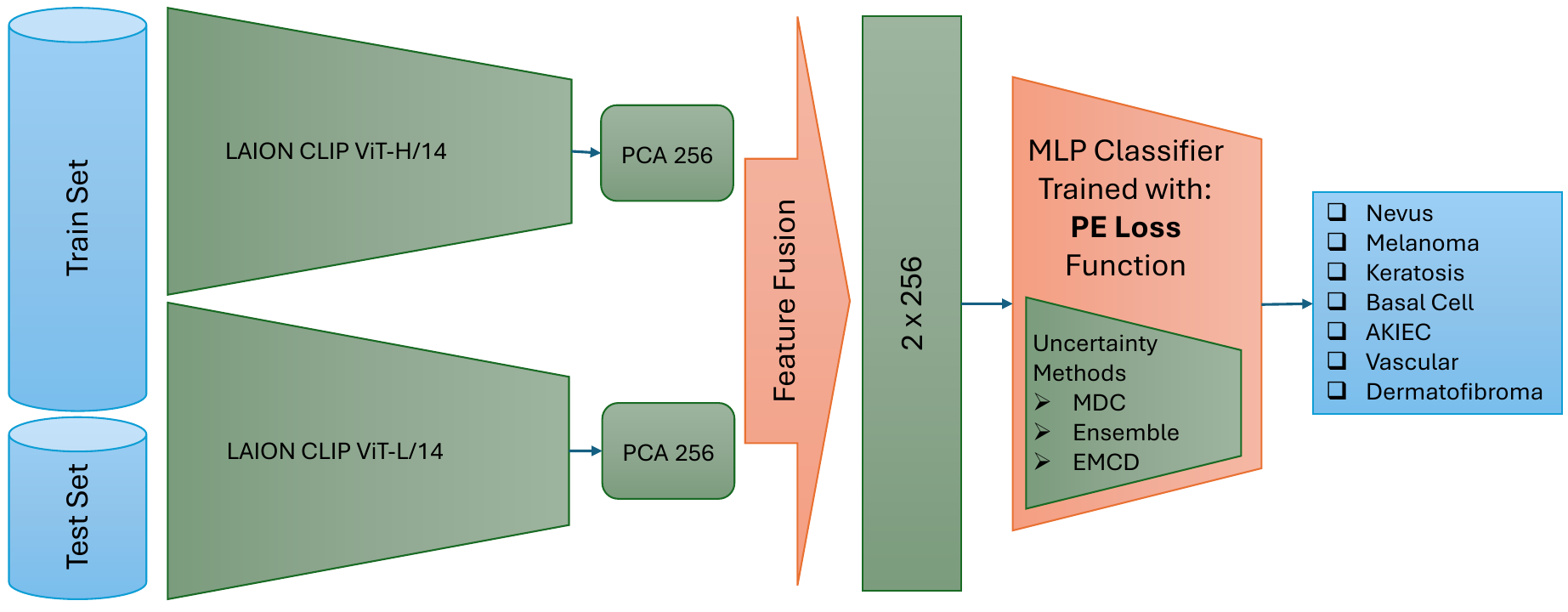} 
    \caption{Architecture of the quad-step proposed method}
    \label{fig:architecture_proposed_method}
\end{figure*}
%------------------------------------------------
\begin{figure*}[t]
  \centering
  \includegraphics[width=0.8\linewidth]{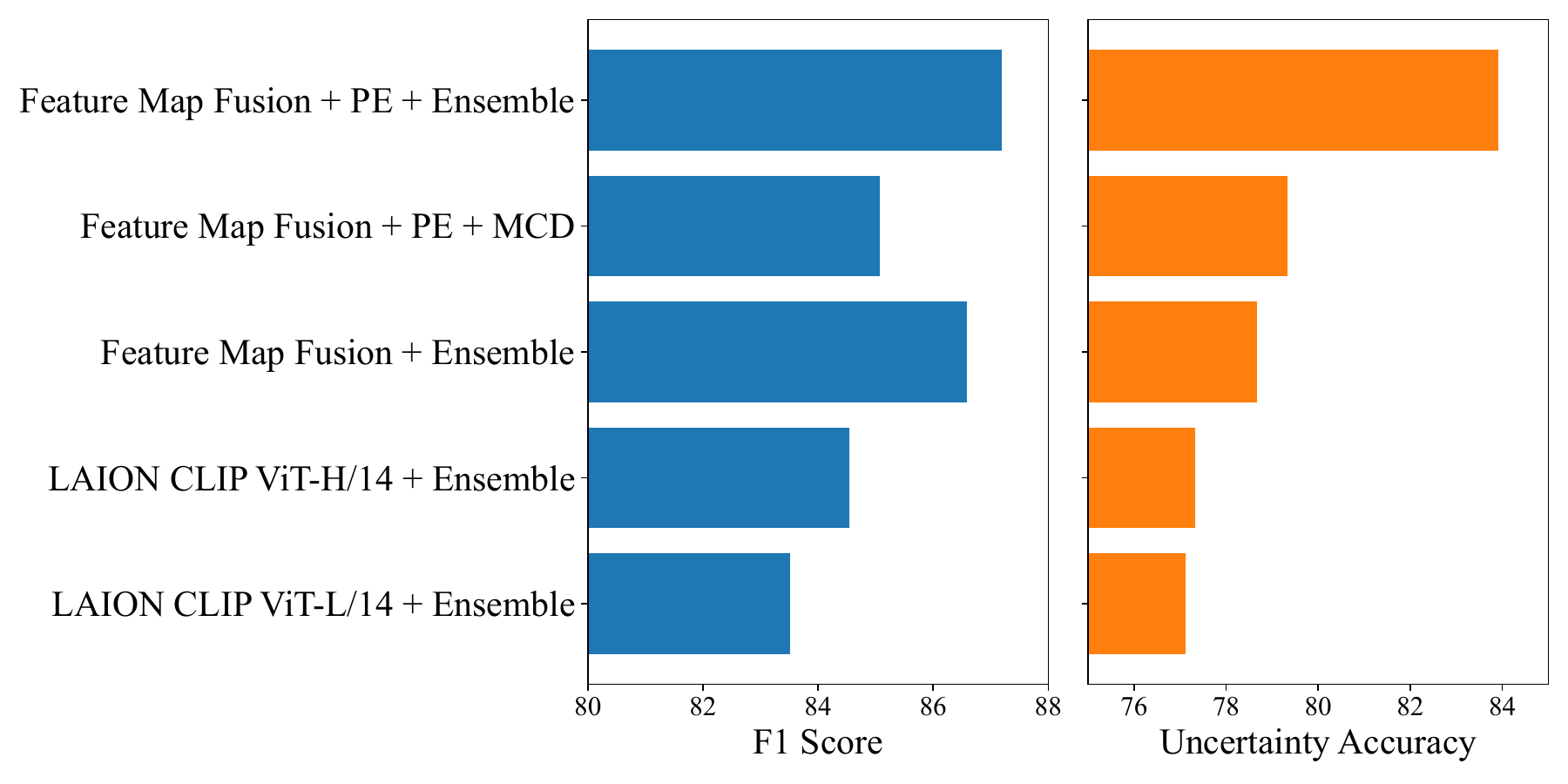}
  \caption{Performance comparison of the best-performing algorithms across various architectures and uncertainty-aware strategies. Evaluated metrics include F1 Score (F1) and Uncertainty Accuracy (UAcc). The Proposed method of this research has the best outcome.}
  \label{fig:uacc_f2}
\end{figure*}
% ------------------------------------------------
\subsection{Discussion}
Taken together, these findings highlight the superiority of CLIP-based vision transformers—particularly ViT-H/14 and ViT-L/14 trained on LAION—over CNN-based feature extractors such as ResNet50, DenseNet121, and VGG16, as well as EfficientNet-V2-L. The advantage of CLIP lies in its training on large-scale image–text pairs, enabling the model to learn semantically rich and globally contextualized features that generalize well to medical images. In contrast, CNNs emphasize local spatial patterns and are less effective at capturing global structures critical for lesion classification. The transformer’s ability to model long-range dependencies further enhances its discriminative power, leading to superior accuracy and more reliable uncertainty calibration.

Our results also demonstrate that margin-based classifiers (SVMs) and tree-based methods (e.g., XGBoost) are well suited to exploit the complex feature spaces generated by CLIP, while simpler models struggle. Moreover, the integration of uncertainty-aware frameworks, particularly Ensemble methods, enhances transparency by providing reliable measures of prediction confidence without sacrificing accuracy. In sensitive domains such as medical imaging, this combination of strong performance and trustworthy uncertainty quantification is essential for building reliable AI-assisted diagnostic systems.
%-------------------------------------------------------------------------------
\subsection{Proposed Method}\label{sec:proposed_method}  
%----------------------------------------------------------------------------
\begin{table*}[t!]
    \caption{Comparing Proposed Method with previous results}
    \label{tab:uncertainty_results_proposed_method} 
    \centering
    \resizebox{0.9\textwidth}{!}{
    \begin{tabular}{llrrrrrrrrrrrr}
        \toprule
        Feature Extractor & Method & Acc & Pre & Recall & F1 & TU & FC & FU & TC & UAcc & USen & USpe & UPre \\
        \midrule
        Feature Map Fusion + PE & Ensemble & 87.62 & 87.16 & 87.62 & 87.19 & 166 & 82 & 240 & 1515 & 83.92 & 66.94 & 86.32 & 40.89 \\
        Feature Map Fusion + PE & MCD & 86.17 & 84.48 & 86.17 & 85.07 & 218 & 59 & 355 & 1371 & 79.33 & 78.70 & 79.43 & 38.05 \\
        Feature Map Fusion & Ensemble & 87.22 & 87.01 & 87.22 & 86.59 & 204 & 52 & 375 & 1372 & 78.68 & 79.69 & 78.53 & 35.23 \\
        LAION CLIP ViT-H/14 & Ensemble & 85.47 & 83.90 & 85.47 & 84.54 & 231 & 60 & 394 & 1318 & 77.33 & 79.38 & 76.99 & 36.96 \\
        LAION CLIP ViT-L/14 & Ensemble & 84.62 & 83.13 & 84.62 & 83.52 & 246 & 62 & 396 & 1299 & 77.13 & 79.87 & 76.64 & 38.32 \\
        Feature Map Fusion + PE & EMCD & 87.62 & 87.16 & 87.62 & 87.18 & 204 & 44 & 424 & 1331 & 76.64 & 82.26 & 75.84 & 32.48 \\
        Feature Map Fusion & MCD & 87.07 & 85.65 & 87.07 & 86.22 & 222 & 37 & 525 & 1219 & 71.94 & 85.71 & 69.90 & 29.72 \\
        Feature Map Fusion & EMCD & 87.12 & 86.90 & 87.12 & 86.45 & 230 & 28 & 538 & 1207 & 71.74 & 89.15 & 69.17 & 29.95 \\
        \bottomrule
    \end{tabular}
    }
\end{table*}
%--------------------------------------------------
Building on the comparative analysis in Section~\ref{sec:results}, where CLIP-based vision transformers (particularly ViT-H/14 and ViT-L/14) combined with PCA 256 achieved the strongest performance, we propose an enhanced framework that integrates the complementary strengths of these top-performing feature extractors. The key intuition is that combining representations from multiple high-capacity models can yield richer, more discriminative features, while predictive entropy (PE) loss further encourages well-calibrated predictions.  

The architecture of the proposed framework, illustrated in Figure~\ref{fig:architecture_proposed_method}, consists of four stages:  

\begin{enumerate}  
    \item \textbf{Feature Extraction:} Input images are processed through two state-of-the-art CLIP-based vision transformers—\textit{LAION CLIP ViT-H/14} and \textit{LAION CLIP ViT-L/14}. Each model generates high-dimensional feature maps that capture complementary aspects of the skin lesion images.  

    \item \textbf{Dimensionality Reduction and Feature Fusion:} The extracted features are reduced to 256 dimensions via \textit{Principal Component Analysis (PCA)}, a configuration shown to be optimal in earlier experiments. The reduced features are then concatenated, forming a joint fused representation ($2 \times 256$) that preserves complementary information from both models.  

    \item \textbf{Classification with Predictive Entropy Loss:} The fused features are passed into a \textit{Multi-Layer Perceptron (MLP)} classifier. Unlike the baseline models, the proposed framework incorporates a \textit{Predictive Entropy (PE)} loss term, which regularizes training by explicitly penalizing overconfident but incorrect predictions. This improves both predictive reliability and uncertainty calibration across the seven lesion categories: Nevus, Melanoma, Keratosis, Basal Cell Carcinoma, AKIEC, Vascular Lesions, and Dermatofibroma.  

    \item \textbf{Uncertainty Quantification:} To provide reliable measures of confidence, three uncertainty-aware approaches—\textit{Monte Carlo Dropout (MCD)}, \textit{Ensemble}, and \textit{Ensemble MCD (EMCD)}—are integrated into the pipeline.  
\end{enumerate}  

Table~\ref{tab:uncertainty_results_proposed_method} compares the proposed method against the best-performing baseline models. Figure~\ref{fig:uacc_f2} illustrates the
performance of the top 5 methods .The results demonstrate that feature fusion combined with predictive entropy (PE) loss consistently outperforms individual models. For instance, with Ensemble uncertainty estimation, the proposed method achieves 87.62\% accuracy, 87.19\% F1, and UAcc = 83.92\%, substantially surpassing the strongest baseline (ViT-H/14 + Ensemble, Acc = 85.47\%, UAcc = 77.33\%). Even without PE loss, fusion models deliver higher accuracy (e.g., Acc = 87.22\%, F1 = 86.59\%) than any single feature extractor.  

These improvements confirm that dual-model feature fusion enriches representation quality by leveraging complementary information, and incorporating PE loss enhances calibration by reducing overconfident errors. Together, these innovations provide a more accurate and trustworthy framework for automated skin lesion classification compared to prior baselines. 
%----------------------------------------------------------------------------

\section{Conclusion}\label{Sec:Conclusion}  

This study conducted a comprehensive evaluation of deep learning-based methods for automated skin cancer classification on the HAM10000 dataset, with a particular focus on both predictive performance and model reliability through uncertainty quantification. CLIP-based vision transformers, especially LAION ViT-H/14 with PCA 256, consistently achieved the strongest baseline performance, highlighting the value of large-scale contrastive pre-training for medical image analysis.  
Motivated by these findings, we proposed a novel framework that combines feature fusion of the two strongest extractors (ViT-H/14 and ViT-L/14) with a Predictive Entropy (PE) loss function. Feature fusion enriched the representation space by leveraging complementary information, while the PE loss improved calibration by discouraging overconfident but incorrect predictions. Together, these enhancements led to state-of-the-art results, with the proposed method achieving 87.62\% accuracy, 87.19\% F1 score, and 83.92\% uncertainty accuracy—substantially outperforming all baseline models.  
In conclusion, the results demonstrate that carefully integrating advanced feature extractors, feature fusion, and uncertainty-aware training objectives provides significant improvements in both accuracy and reliability. This combination offers a promising direction for the development of transparent and trustworthy AI systems to support clinical decision-making in dermatology.  
%---------------------------------------------------
\section{Acknowledgment}

%---------------------------------------------------

\bibliographystyle{elsarticle-num} 
\bibliography{main.bib}

%---------------------------------------------------
\appendix

%---------------------------------------------------
\clearpage
\onecolumn
\section{Results Tables} \label{app:ham10000_results}

Table~\ref{tab:performance_metrics_by_feature_extractor} presents a detailed comparison of classification performance for all evaluated methods and feature extractors on the HAM10000 dataset.  
The table expands upon the main results by including all intermediate results and by reporting additional performance measures not listed in the main text.  
The metric definitions are the same as those described in Section~\ref{sec:results}.

\input{long_table_1}

%---------------------------------------------------

\input{long_table_2}

\end{document}

%% file: long_table_1.tex
\small
\begin{longtable}{llccccc}
    \caption{Comparison of classification performance across various methods and feature extractors on the HAM10000 dataset.}
    \label{tab:performance_metrics_by_feature_extractor} \\
    \hline
    \textbf{Feature Extractor} & \textbf{PCA NC} & \textbf{Method} & \textbf{Acc} & \textbf{Pre} & \textbf{Recall} & \textbf{F1} \\
    \hline
    \endfirsthead
    \hline
    \textbf{Feature Extractor} & \textbf{PCA NC} & \textbf{Method} & \textbf{Acc} & \textbf{Pre} & \textbf{Recall} & \textbf{F1} \\
    \hline
    \endhead
    \hline
    \endfoot    
    LAION CLIP ViT-H/14 & 512 & SVM & 85.54 & 84.81 & 85.54 & 84.41 \\
    LAION CLIP ViT-H/14 & 256 & SVM & 85.39 & 84.66 & 85.39 & 84.27 \\
    LAION CLIP ViT-H/14 & 512 & LogisticRegression & 84.48 & 84.04 & 84.48 & 84.17 \\
    LAION CLIP ViT-L/14 & 512 & SVM & 85.10 & 84.28 & 85.10 & 83.98 \\
    LAION CLIP ViT-L/14 & 256 & SVM & 84.98 & 84.15 & 84.98 & 83.85 \\
    LAION CLIP ViT-H/14 & 256 & LogisticRegression & 84.18 & 83.55 & 84.18 & 83.74 \\
    LAION CLIP ViT-H/14 & 128 & SVM & 84.82 & 84.07 & 84.82 & 83.66 \\
    LAION CLIP ViT-L/14 & 128 & SVM & 84.71 & 83.87 & 84.71 & 83.57 \\
    LAION CLIP ViT-L/14 & 512 & LogisticRegression & 83.17 & 82.75 & 83.17 & 82.89 \\
    LAION CLIP ViT-L/14 & 256 & LogisticRegression & 83.21 & 82.51 & 83.21 & 82.74 \\
    LAION CLIP ViT-L/14 & 64 & SVM & 83.82 & 82.86 & 83.82 & 82.54 \\
    LAION CLIP ViT-H/14 & 64 & SVM & 83.81 & 82.89 & 83.81 & 82.51 \\
    OpenAI CLIP ViT-L/14 & 512 & SVM & 83.50 & 82.63 & 83.50 & 82.19 \\
    OpenAI CLIP ViT-L/14 & 256 & SVM & 83.26 & 82.35 & 83.26 & 81.94 \\
    LAION CLIP ViT-H/14 & 128 & LogisticRegression & 82.53 & 81.59 & 82.53 & 81.88 \\
    LAION CLIP ViT-L/14 & 128 & LogisticRegression & 82.42 & 81.51 & 82.42 & 81.80 \\
    OpenAI CLIP ViT-B/16 & 512 & LogisticRegression & 82.48 & 81.51 & 82.48 & 81.72 \\
    OpenAI CLIP ViT-L/14 & 512 & LogisticRegression & 82.15 & 81.51 & 82.15 & 81.72 \\
    OpenAI CLIP ViT-L/14 & 128 & SVM & 82.68 & 81.66 & 82.68 & 81.28 \\
    OpenAI CLIP ViT-L/14 & 256 & LogisticRegression & 81.84 & 80.94 & 81.84 & 81.24 \\
    OpenAI CLIP ViT-B/32 & 512 & LogisticRegression & 81.91 & 80.93 & 81.91 & 81.15 \\
    OpenAI CLIP ViT-B/16 & 512 & SVM & 82.60 & 80.57 & 82.60 & 80.97 \\
    OpenAI CLIP ViT-B/16 & 256 & LogisticRegression & 81.70 & 80.69 & 81.70 & 80.93 \\
    OpenAI CLIP ViT-B/16 & 256 & SVM & 82.52 & 80.49 & 82.52 & 80.89 \\
    OpenAI CLIP ViT-B/32 & 256 & LogisticRegression & 81.36 & 80.25 & 81.36 & 80.54 \\
    OpenAI CLIP ViT-B/16 & 128 & SVM & 82.00 & 79.89 & 82.00 & 80.34 \\
    OpenAI CLIP ViT-L/14 & 128 & LogisticRegression & 80.90 & 79.74 & 80.90 & 80.11 \\
    LAION CLIP ViT-H/14 & 64 & LogisticRegression & 80.94 & 79.64 & 80.94 & 80.01 \\
    LAION CLIP ViT-L/14 & 64 & LogisticRegression & 80.96 & 79.63 & 80.96 & 79.99 \\
    OpenAI CLIP ViT-B/32 & 512 & SVM & 81.66 & 80.05 & 81.66 & 79.91 \\
    OpenAI CLIP ViT-L/14 & 64 & SVM & 81.38 & 80.20 & 81.38 & 79.87 \\
    OpenAI CLIP ViT-B/32 & 256 & SVM & 81.56 & 80.04 & 81.56 & 79.81 \\
    LAION CLIP ViT-H/14 & 64 & Xgboost & 81.26 & 79.81 & 81.26 & 79.53 \\
    LAION CLIP ViT-H/14 & 128 & Xgboost & 81.42 & 79.91 & 81.42 & 79.46 \\
    OpenAI CLIP ViT-B/32 & 128 & SVM & 81.07 & 79.37 & 81.07 & 79.29 \\
    OpenAI CLIP ViT-B/16 & 128 & LogisticRegression & 80.22 & 78.92 & 80.22 & 79.22 \\
    OpenAI CLIP ViT-B/16 & 64 & SVM & 81.00 & 78.71 & 81.00 & 79.19 \\
    ResNet50 & 512 & LogisticRegression & 79.68 & 78.82 & 79.68 & 79.13 \\
    LAION CLIP ViT-L/14 & 64 & Xgboost & 80.86 & 79.36 & 80.86 & 79.12 \\
    LAION CLIP ViT-H/14 & 256 & Xgboost & 81.15 & 79.85 & 81.15 & 79.00 \\
    LAION CLIP ViT-L/14 & 128 & Xgboost & 80.89 & 79.48 & 80.89 & 78.97 \\
    LAION CLIP ViT-L/14 & 256 & Xgboost & 80.84 & 79.43 & 80.84 & 78.72 \\
    OpenAI CLIP ViT-L/14 & 64 & LogisticRegression & 79.67 & 78.27 & 79.67 & 78.68 \\
    OpenAI CLIP ViT-B/32 & 128 & LogisticRegression & 79.61 & 78.22 & 79.61 & 78.57 \\
    OpenAI CLIP ViT-B/32 & 64 & SVM & 80.39 & 78.34 & 80.39 & 78.52 \\
    LAION CLIP ViT-H/14 & 512 & Xgboost & 80.69 & 79.11 & 80.69 & 78.31 \\
    ResNet50 & 256 & LogisticRegression & 78.90 & 77.76 & 78.90 & 78.12 \\
    LAION CLIP ViT-H/14 & 512 & KNeighbors & 79.24 & 78.16 & 79.24 & 78.10 \\
    LAION CLIP ViT-H/14 & 256 & KNeighbors & 79.17 & 77.94 & 79.17 & 78.00 \\
    LAION CLIP ViT-L/14 & 128 & KNeighbors & 79.04 & 77.87 & 79.04 & 77.92 \\
    LAION CLIP ViT-L/14 & 256 & KNeighbors & 79.06 & 77.74 & 79.06 & 77.84 \\
    LAION CLIP ViT-H/14 & 128 & KNeighbors & 78.80 & 77.61 & 78.80 & 77.70 \\
    LAION CLIP ViT-L/14 & 512 & KNeighbors & 78.93 & 77.58 & 78.93 & 77.70 \\
    LAION CLIP ViT-L/14 & 64 & KNeighbors & 78.67 & 77.49 & 78.67 & 77.61 \\
    LAION CLIP ViT-L/14 & 512 & Xgboost & 79.85 & 78.19 & 79.85 & 77.32 \\
    OpenAI CLIP ViT-B/16 & 64 & LogisticRegression & 78.57 & 76.94 & 78.57 & 77.30 \\
    LAION CLIP ViT-H/14 & 64 & KNeighbors & 78.40 & 77.11 & 78.40 & 77.24 \\
    ResNet50 & 512 & SVM & 79.32 & 77.14 & 79.32 & 77.17 \\
    DenseNet121 & 512 & SVM & 79.44 & 76.97 & 79.44 & 77.03 \\
    ResNet50 & 256 & SVM & 79.15 & 76.95 & 79.15 & 77.00 \\
    OpenAI CLIP ViT-B/32 & 64 & LogisticRegression & 78.38 & 76.42 & 78.38 & 76.96 \\
    OpenAI CLIP ViT-L/14 & 64 & Xgboost & 78.93 & 77.06 & 78.93 & 76.89 \\
    DenseNet121 & 256 & SVM & 79.15 & 76.68 & 79.15 & 76.76 \\
    OpenAI CLIP ViT-L/14 & 128 & Xgboost & 79.08 & 77.09 & 79.08 & 76.73 \\
    ResNet50 & 128 & SVM & 78.71 & 76.40 & 78.71 & 76.49 \\
    OpenAI CLIP ViT-L/14 & 256 & KNeighbors & 77.35 & 76.16 & 77.35 & 76.17 \\
    DenseNet121 & 128 & SVM & 78.53 & 75.87 & 78.53 & 76.09 \\
    OpenAI CLIP ViT-L/14 & 256 & Xgboost & 78.73 & 76.66 & 78.73 & 76.08 \\
    ResNet50 & 128 & LogisticRegression & 77.31 & 75.44 & 77.31 & 75.97 \\
    OpenAI CLIP ViT-L/14 & 512 & KNeighbors & 77.21 & 75.97 & 77.21 & 75.97 \\
    OpenAI CLIP ViT-L/14 & 128 & KNeighbors & 77.11 & 75.80 & 77.11 & 75.95 \\
    DenseNet121 & 512 & LogisticRegression & 76.23 & 75.81 & 76.23 & 75.91 \\
    DenseNet121 & 256 & LogisticRegression & 76.65 & 75.21 & 76.65 & 75.72 \\
    ResNet50 & 64 & SVM & 78.00 & 75.26 & 78.00 & 75.61 \\
    OpenAI CLIP ViT-B/16 & 128 & Xgboost & 78.32 & 75.42 & 78.32 & 75.48 \\
    OpenAI CLIP ViT-B/16 & 64 & Xgboost & 78.07 & 75.22 & 78.07 & 75.48 \\
    OpenAI CLIP ViT-B/32 & 64 & Xgboost & 77.89 & 75.73 & 77.89 & 75.46 \\
    OpenAI CLIP ViT-B/32 & 128 & Xgboost & 77.94 & 75.57 & 77.94 & 75.25 \\
    OpenAI CLIP ViT-L/14 & 64 & KNeighbors & 76.14 & 74.84 & 76.14 & 74.94 \\
    OpenAI CLIP ViT-L/14 & 512 & Xgboost & 77.82 & 75.41 & 77.82 & 74.80 \\
    OpenAI CLIP ViT-B/32 & 256 & Xgboost & 77.80 & 75.41 & 77.80 & 74.79 \\
    DenseNet121 & 64 & SVM & 77.42 & 74.13 & 77.42 & 74.77 \\
    VGG16 & 256 & LogisticRegression & 75.72 & 74.03 & 75.72 & 74.58 \\
    DenseNet121 & 128 & LogisticRegression & 76.04 & 73.93 & 76.04 & 74.54 \\
    OpenAI CLIP ViT-B/16 & 256 & Xgboost & 77.68 & 74.84 & 77.68 & 74.46 \\
    LAION CLIP ViT-L/14 & 64 & NaiveBayes & 75.30 & 74.30 & 75.30 & 74.39 \\
    VGG16 & 256 & SVM & 77.44 & 74.61 & 77.44 & 74.34 \\
    VGG16 & 512 & SVM & 77.37 & 74.70 & 77.37 & 74.18 \\
    LAION CLIP ViT-H/14 & 64 & NaiveBayes & 74.92 & 74.11 & 74.92 & 74.08 \\
    VGG16 & 128 & SVM & 77.17 & 74.15 & 77.17 & 74.03 \\
    VGG16 & 512 & LogisticRegression & 74.15 & 73.78 & 74.15 & 73.87 \\
    ResNet50 & 64 & LogisticRegression & 75.54 & 73.10 & 75.54 & 73.76 \\
    OpenAI CLIP ViT-B/16 & 256 & KNeighbors & 75.35 & 73.62 & 75.35 & 73.72 \\
    OpenAI CLIP ViT-B/16 & 512 & KNeighbors & 75.38 & 73.58 & 75.38 & 73.70 \\
    ResNet50 & 64 & Xgboost & 76.35 & 73.81 & 76.35 & 73.48 \\
    OpenAI CLIP ViT-B/16 & 512 & Xgboost & 77.03 & 73.90 & 77.03 & 73.43 \\
    OpenAI CLIP ViT-B/32 & 512 & Xgboost & 76.83 & 74.16 & 76.83 & 73.38 \\
    OpenAI CLIP ViT-B/16 & 128 & KNeighbors & 74.97 & 73.03 & 74.97 & 73.32 \\
    VGG16 & 64 & SVM & 76.29 & 73.12 & 76.29 & 73.00 \\
    ResNet50 & 128 & Xgboost & 76.12 & 73.33 & 76.12 & 73.00 \\
    VGG16 & 128 & LogisticRegression & 74.79 & 72.17 & 74.79 & 72.95 \\
    OpenAI CLIP ViT-B/16 & 64 & KNeighbors & 74.54 & 72.59 & 74.54 & 72.91 \\
    OpenAI CLIP ViT-B/32 & 256 & KNeighbors & 74.74 & 72.51 & 74.74 & 72.77 \\
    OpenAI CLIP ViT-B/32 & 512 & KNeighbors & 74.75 & 72.46 & 74.75 & 72.76 \\
    OpenAI CLIP ViT-B/32 & 128 & KNeighbors & 74.61 & 72.50 & 74.61 & 72.68 \\
    DenseNet121 & 64 & LogisticRegression & 74.73 & 71.84 & 74.73 & 72.64 \\
    ResNet50 & 256 & Xgboost & 76.03 & 73.26 & 76.03 & 72.48 \\
    DenseNet121 & 64 & Xgboost & 75.51 & 72.30 & 75.51 & 72.46 \\
    ResNet50 & 256 & KNeighbors & 73.70 & 72.35 & 73.70 & 72.31 \\
    DenseNet121 & 128 & Xgboost & 75.64 & 72.56 & 75.64 & 72.27 \\
    ResNet50 & 128 & KNeighbors & 73.52 & 72.10 & 73.52 & 72.14 \\
    ResNet50 & 512 & KNeighbors & 73.56 & 72.15 & 73.56 & 72.13 \\
    OpenAI CLIP ViT-B/32 & 64 & KNeighbors & 73.96 & 71.56 & 73.96 & 72.06 \\
    LAION CLIP ViT-L/14 & 128 & NaiveBayes & 72.60 & 72.52 & 72.60 & 72.05 \\
    ResNet50 & 64 & KNeighbors & 73.41 & 71.67 & 73.41 & 72.01 \\
    LAION CLIP ViT-H/14 & 128 & NaiveBayes & 72.18 & 72.75 & 72.18 & 71.83 \\
    OpenAI CLIP ViT-B/32 & 64 & NaiveBayes & 73.09 & 71.08 & 73.09 & 71.64 \\
    DenseNet121 & 256 & Xgboost & 75.27 & 72.17 & 75.27 & 71.58 \\
    OpenAI CLIP ViT-B/16 & 64 & NaiveBayes & 73.02 & 71.13 & 73.02 & 71.53 \\
    DenseNet121 & 256 & KNeighbors & 73.20 & 71.01 & 73.20 & 71.29 \\
    DenseNet121 & 128 & KNeighbors & 72.96 & 71.01 & 72.96 & 71.26 \\
    ResNet50 & 512 & Xgboost & 75.27 & 72.45 & 75.27 & 71.20 \\
    VGG16 & 64 & Xgboost & 74.48 & 70.91 & 74.48 & 71.02 \\
    VGG16 & 64 & LogisticRegression & 73.59 & 70.09 & 73.59 & 70.92 \\
    VGG16 & 128 & Xgboost & 74.72 & 71.37 & 74.72 & 70.91 \\
    DenseNet121 & 512 & KNeighbors & 72.97 & 70.71 & 72.97 & 70.77 \\
    DenseNet121 & 512 & Xgboost & 74.80 & 71.29 & 74.80 & 70.52 \\
    DenseNet121 & 64 & KNeighbors & 71.79 & 69.87 & 71.79 & 70.23 \\
    VGG16 & 256 & Xgboost & 74.42 & 71.18 & 74.42 & 70.17 \\
    OpenAI CLIP ViT-L/14 & 64 & NaiveBayes & 70.54 & 70.17 & 70.54 & 69.88 \\
    VGG16 & 128 & KNeighbors & 72.29 & 68.79 & 72.29 & 69.52 \\
    VGG16 & 64 & KNeighbors & 71.89 & 68.67 & 71.89 & 69.51 \\
    VGG16 & 256 & KNeighbors & 72.32 & 68.59 & 72.32 & 69.20 \\
    OpenAI CLIP ViT-B/16 & 128 & NaiveBayes & 69.81 & 69.94 & 69.81 & 69.12 \\
    VGG16 & 512 & Xgboost & 73.83 & 70.23 & 73.83 & 69.09 \\
    LAION CLIP ViT-L/14 & 64 & RandomForest & 74.66 & 72.47 & 74.66 & 68.92 \\
    VGG16 & 512 & KNeighbors & 72.31 & 68.49 & 72.31 & 68.92 \\
    LAION CLIP ViT-H/14 & 64 & RandomForest & 74.56 & 72.77 & 74.56 & 68.59 \\
    EfficientNet-V2-L & 512 & LogisticRegression & 69.98 & 67.56 & 69.98 & 68.49 \\
    OpenAI CLIP ViT-B/32 & 128 & NaiveBayes & 69.23 & 68.70 & 69.23 & 68.43 \\
    EfficientNet-V2-L & 256 & LogisticRegression & 70.82 & 67.14 & 70.82 & 68.30 \\
    LAION CLIP ViT-L/14 & 256 & NaiveBayes & 68.33 & 69.58 & 68.33 & 68.02 \\
    LAION CLIP ViT-H/14 & 256 & NaiveBayes & 67.69 & 69.49 & 67.69 & 67.58 \\
    OpenAI CLIP ViT-L/14 & 128 & NaiveBayes & 67.70 & 68.23 & 67.70 & 67.42 \\
    EfficientNet-V2-L & 128 & LogisticRegression & 70.68 & 65.81 & 70.68 & 66.86 \\
    LAION CLIP ViT-H/14 & 64 & GradientBoosting & 72.50 & 68.01 & 72.50 & 66.43 \\
    ResNet50 & 64 & NaiveBayes & 67.37 & 66.27 & 67.37 & 66.42 \\
    EfficientNet-V2-L & 512 & KNeighbors & 68.62 & 64.72 & 68.62 & 66.02 \\
    LAION CLIP ViT-L/14 & 64 & GradientBoosting & 72.12 & 68.38 & 72.12 & 65.98 \\
    EfficientNet-V2-L & 64 & Xgboost & 71.31 & 65.06 & 71.31 & 65.94 \\
    LAION CLIP ViT-H/14 & 256 & GradientBoosting & 72.15 & 67.58 & 72.15 & 65.93 \\
    EfficientNet-V2-L & 128 & Xgboost & 71.77 & 65.75 & 71.77 & 65.88 \\
    LAION CLIP ViT-H/14 & 128 & GradientBoosting & 72.06 & 67.44 & 72.06 & 65.86 \\
    EfficientNet-V2-L & 256 & KNeighbors & 68.40 & 64.51 & 68.40 & 65.82 \\
    LAION CLIP ViT-L/14 & 64 & DecisionTree & 65.72 & 65.95 & 65.72 & 65.81 \\
    LAION CLIP ViT-L/14 & 128 & GradientBoosting & 71.87 & 67.81 & 71.87 & 65.65 \\
    LAION CLIP ViT-L/14 & 256 & GradientBoosting & 71.87 & 67.78 & 71.87 & 65.61 \\
    DenseNet121 & 64 & NaiveBayes & 65.46 & 67.15 & 65.46 & 65.60 \\
    LAION CLIP ViT-H/14 & 512 & GradientBoosting & 71.83 & 67.13 & 71.83 & 65.58 \\
    LAION CLIP ViT-L/14 & 512 & GradientBoosting & 71.63 & 67.55 & 71.63 & 65.50 \\
    OpenAI CLIP ViT-L/14 & 64 & RandomForest & 72.35 & 68.84 & 72.35 & 65.41 \\
    EfficientNet-V2-L & 128 & KNeighbors & 67.86 & 63.88 & 67.86 & 65.37 \\
    EfficientNet-V2-L & 512 & SVM & 72.11 & 66.55 & 72.11 & 65.28 \\
    LAION CLIP ViT-H/14 & 64 & DecisionTree & 65.15 & 65.34 & 65.15 & 65.22 \\
    EfficientNet-V2-L & 64 & LogisticRegression & 70.16 & 63.86 & 70.16 & 65.15 \\
    EfficientNet-V2-L & 64 & KNeighbors & 67.69 & 63.59 & 67.69 & 65.08 \\
    EfficientNet-V2-L & 256 & Xgboost & 71.56 & 65.34 & 71.56 & 65.07 \\
    EfficientNet-V2-L & 256 & SVM & 71.89 & 66.06 & 71.89 & 65.03 \\
    OpenAI CLIP ViT-B/16 & 256 & NaiveBayes & 65.21 & 67.86 & 65.21 & 64.96 \\
    LAION CLIP ViT-L/14 & 128 & DecisionTree & 64.85 & 65.09 & 64.85 & 64.95 \\
    OpenAI CLIP ViT-B/32 & 64 & GradientBoosting & 71.16 & 65.14 & 71.16 & 64.87 \\
    OpenAI CLIP ViT-L/14 & 256 & NaiveBayes & 64.83 & 66.19 & 64.83 & 64.86 \\
    LAION CLIP ViT-H/14 & 128 & DecisionTree & 64.62 & 65.12 & 64.62 & 64.85 \\
    OpenAI CLIP ViT-B/32 & 128 & GradientBoosting & 71.06 & 64.70 & 71.06 & 64.60 \\
    EfficientNet-V2-L & 128 & SVM & 71.51 & 65.74 & 71.51 & 64.40 \\
    OpenAI CLIP ViT-B/32 & 256 & GradientBoosting & 70.83 & 64.54 & 70.83 & 64.39 \\
    DenseNet121 & 64 & RandomForest & 71.38 & 67.14 & 71.38 & 64.29 \\
    DenseNet121 & 128 & NaiveBayes & 62.94 & 67.88 & 62.94 & 64.21 \\
    LAION CLIP ViT-L/14 & 512 & NaiveBayes & 64.37 & 66.39 & 64.37 & 64.14 \\
    LAION CLIP ViT-H/14 & 512 & NaiveBayes & 64.09 & 66.68 & 64.09 & 64.14 \\
    OpenAI CLIP ViT-B/32 & 256 & NaiveBayes & 64.43 & 65.53 & 64.43 & 63.93 \\
    OpenAI CLIP ViT-L/14 & 64 & DecisionTree & 63.83 & 63.83 & 63.83 & 63.81 \\
    LAION CLIP ViT-H/14 & 128 & RandomForest & 71.83 & 71.48 & 71.83 & 63.79 \\
    ResNet50 & 128 & NaiveBayes & 63.88 & 64.69 & 63.88 & 63.78 \\
    LAION CLIP ViT-H/14 & 256 & DecisionTree & 63.79 & 63.81 & 63.79 & 63.78 \\
    LAION CLIP ViT-L/14 & 128 & RandomForest & 71.80 & 70.65 & 71.80 & 63.78 \\
    LAION CLIP ViT-L/14 & 256 & DecisionTree & 63.52 & 63.90 & 63.52 & 63.69 \\
    OpenAI CLIP ViT-B/32 & 64 & RandomForest & 71.42 & 68.54 & 71.42 & 63.61 \\
    EfficientNet-V2-L & 512 & Xgboost & 70.99 & 64.00 & 70.99 & 63.53 \\
    DenseNet121 & 64 & GradientBoosting & 70.09 & 64.59 & 70.09 & 63.42 \\
    LAION CLIP ViT-H/14 & 512 & DecisionTree & 63.15 & 63.72 & 63.15 & 63.40 \\
    ResNet50 & 64 & RandomForest & 71.09 & 65.96 & 71.09 & 63.31 \\
    LAION CLIP ViT-L/14 & 512 & DecisionTree & 63.26 & 63.40 & 63.26 & 63.31 \\
    OpenAI CLIP ViT-B/16 & 64 & GradientBoosting & 70.66 & 64.16 & 70.66 & 63.29 \\
    OpenAI CLIP ViT-L/14 & 64 & GradientBoosting & 70.71 & 65.10 & 70.71 & 63.28 \\
    DenseNet121 & 128 & GradientBoosting & 69.95 & 63.65 & 69.95 & 63.25 \\
    EfficientNet-V2-L & 64 & SVM & 70.94 & 64.28 & 70.94 & 63.18 \\
    OpenAI CLIP ViT-B/16 & 128 & GradientBoosting & 70.52 & 64.05 & 70.52 & 63.14 \\
    OpenAI CLIP ViT-L/14 & 128 & DecisionTree & 62.95 & 63.27 & 62.95 & 63.09 \\
    OpenAI CLIP ViT-B/32 & 512 & GradientBoosting & 70.13 & 62.95 & 70.13 & 63.08 \\
    DenseNet121 & 256 & GradientBoosting & 69.89 & 63.75 & 69.89 & 63.01 \\
    VGG16 & 64 & RandomForest & 70.69 & 65.88 & 70.69 & 63.00 \\
    OpenAI CLIP ViT-B/16 & 64 & RandomForest & 71.07 & 65.81 & 71.07 & 62.99 \\
    OpenAI CLIP ViT-B/16 & 512 & GradientBoosting & 70.42 & 63.88 & 70.42 & 62.95 \\
    OpenAI CLIP ViT-B/16 & 256 & GradientBoosting & 70.35 & 63.95 & 70.35 & 62.94 \\
    OpenAI CLIP ViT-L/14 & 128 & GradientBoosting & 70.51 & 64.94 & 70.51 & 62.92 \\
    OpenAI CLIP ViT-B/16 & 64 & DecisionTree & 62.69 & 62.98 & 62.69 & 62.81 \\
    OpenAI CLIP ViT-L/14 & 256 & GradientBoosting & 70.47 & 64.64 & 70.47 & 62.73 \\
    DenseNet121 & 512 & GradientBoosting & 69.60 & 62.92 & 69.60 & 62.72 \\
    OpenAI CLIP ViT-B/32 & 64 & DecisionTree & 62.37 & 62.61 & 62.37 & 62.48 \\
    OpenAI CLIP ViT-B/16 & 128 & DecisionTree & 62.25 & 62.75 & 62.25 & 62.48 \\
    OpenAI CLIP ViT-L/14 & 512 & NaiveBayes & 61.95 & 64.34 & 61.95 & 62.46 \\
    OpenAI CLIP ViT-L/14 & 512 & GradientBoosting & 70.16 & 64.60 & 70.16 & 62.43 \\
    OpenAI CLIP ViT-L/14 & 256 & DecisionTree & 62.15 & 62.57 & 62.15 & 62.34 \\
    ResNet50 & 128 & GradientBoosting & 69.87 & 64.00 & 69.87 & 62.09 \\
    ResNet50 & 64 & GradientBoosting & 69.86 & 63.72 & 69.86 & 62.03 \\
    ResNet50 & 64 & DecisionTree & 61.75 & 62.33 & 61.75 & 62.02 \\
    DenseNet121 & 64 & DecisionTree & 61.83 & 62.24 & 61.83 & 62.02 \\
    VGG16 & 64 & NaiveBayes & 63.71 & 61.22 & 63.71 & 61.97 \\
    OpenAI CLIP ViT-B/16 & 256 & DecisionTree & 61.70 & 62.16 & 61.70 & 61.90 \\
    OpenAI CLIP ViT-B/32 & 128 & DecisionTree & 61.70 & 62.07 & 61.70 & 61.87 \\
    OpenAI CLIP ViT-L/14 & 512 & DecisionTree & 61.69 & 62.04 & 61.69 & 61.84 \\
    ResNet50 & 256 & NaiveBayes & 61.57 & 63.48 & 61.57 & 61.82 \\
    EfficientNet-V2-L & 64 & RandomForest & 70.21 & 63.34 & 70.21 & 61.75 \\
    ResNet50 & 512 & GradientBoosting & 69.62 & 63.14 & 69.62 & 61.74 \\
    ResNet50 & 256 & GradientBoosting & 69.57 & 62.93 & 69.57 & 61.59 \\
    DenseNet121 & 256 & NaiveBayes & 59.75 & 66.75 & 59.75 & 61.54 \\
    OpenAI CLIP ViT-L/14 & 128 & RandomForest & 70.24 & 66.32 & 70.24 & 61.20 \\
    OpenAI CLIP ViT-B/32 & 256 & DecisionTree & 61.07 & 61.34 & 61.07 & 61.19 \\
    DenseNet121 & 128 & RandomForest & 70.06 & 65.79 & 70.06 & 61.13 \\
    ResNet50 & 128 & DecisionTree & 60.89 & 61.37 & 60.89 & 61.11 \\
    OpenAI CLIP ViT-B/16 & 512 & NaiveBayes & 61.48 & 65.24 & 61.48 & 61.08 \\
    DenseNet121 & 128 & DecisionTree & 60.79 & 61.44 & 60.79 & 61.08 \\
    VGG16 & 64 & GradientBoosting & 69.13 & 62.44 & 69.13 & 61.02 \\
    OpenAI CLIP ViT-B/16 & 512 & DecisionTree & 60.85 & 61.21 & 60.85 & 61.01 \\
    DenseNet121 & 256 & DecisionTree & 60.70 & 61.27 & 60.70 & 60.96 \\
    VGG16 & 128 & GradientBoosting & 69.06 & 61.94 & 69.06 & 60.93 \\
    VGG16 & 64 & DecisionTree & 60.56 & 61.15 & 60.56 & 60.82 \\
    ResNet50 & 256 & DecisionTree & 60.37 & 61.19 & 60.37 & 60.75 \\
    VGG16 & 512 & GradientBoosting & 68.99 & 61.46 & 68.99 & 60.74 \\
    VGG16 & 256 & GradientBoosting & 68.86 & 61.13 & 68.86 & 60.74 \\
    OpenAI CLIP ViT-B/32 & 512 & DecisionTree & 60.60 & 60.78 & 60.60 & 60.68 \\
    DenseNet121 & 512 & DecisionTree & 60.03 & 60.42 & 60.03 & 60.21 \\
    EfficientNet-V2-L & 128 & RandomForest & 69.63 & 62.84 & 69.63 & 60.15 \\
    OpenAI CLIP ViT-B/32 & 512 & NaiveBayes & 60.33 & 63.40 & 60.33 & 60.14 \\
    VGG16 & 128 & RandomForest & 69.39 & 64.40 & 69.39 & 60.03 \\
    ResNet50 & 512 & DecisionTree & 59.72 & 60.30 & 59.72 & 59.99 \\
    ResNet50 & 128 & RandomForest & 69.52 & 64.93 & 69.52 & 59.95 \\
    VGG16 & 128 & NaiveBayes & 61.15 & 60.08 & 61.15 & 59.92 \\
    EfficientNet-V2-L & 128 & GradientBoosting & 68.78 & 60.90 & 68.78 & 59.54 \\
    EfficientNet-V2-L & 64 & GradientBoosting & 68.68 & 61.12 & 68.68 & 59.50 \\
    VGG16 & 256 & DecisionTree & 59.22 & 59.73 & 59.22 & 59.45 \\
    VGG16 & 128 & DecisionTree & 59.07 & 59.83 & 59.07 & 59.43 \\
    OpenAI CLIP ViT-B/32 & 128 & RandomForest & 69.42 & 67.05 & 69.42 & 59.35 \\
    EfficientNet-V2-L & 512 & GradientBoosting & 68.63 & 58.27 & 68.63 & 59.32 \\
    EfficientNet-V2-L & 256 & GradientBoosting & 68.65 & 60.52 & 68.65 & 59.31 \\
    OpenAI CLIP ViT-B/16 & 128 & RandomForest & 69.30 & 64.69 & 69.30 & 59.10 \\
    ResNet50 & 512 & NaiveBayes & 58.43 & 61.74 & 58.43 & 58.90 \\
    VGG16 & 512 & DecisionTree & 58.37 & 59.33 & 58.37 & 58.82 \\
    LAION CLIP ViT-L/14 & 256 & RandomForest & 69.18 & 67.80 & 69.18 & 58.75 \\
    LAION CLIP ViT-H/14 & 256 & RandomForest & 69.22 & 68.85 & 69.22 & 58.75 \\
    DenseNet121 & 512 & NaiveBayes & 57.06 & 63.76 & 57.06 & 58.50 \\
    EfficientNet-V2-L & 256 & RandomForest & 68.90 & 59.03 & 68.90 & 58.45 \\
    DenseNet121 & 256 & RandomForest & 68.60 & 64.75 & 68.60 & 57.80 \\
    OpenAI CLIP ViT-L/14 & 256 & RandomForest & 68.54 & 62.93 & 68.54 & 57.65 \\
    VGG16 & 256 & RandomForest & 68.42 & 63.34 & 68.42 & 57.56 \\
    EfficientNet-V2-L & 64 & DecisionTree & 57.16 & 57.87 & 57.16 & 57.49 \\
    EfficientNet-V2-L & 512 & RandomForest & 68.42 & 57.03 & 68.42 & 57.36 \\
    EfficientNet-V2-L & 256 & DecisionTree & 56.41 & 57.45 & 56.41 & 56.90 \\
    VGG16 & 256 & NaiveBayes & 58.01 & 57.89 & 58.01 & 56.87 \\
    EfficientNet-V2-L & 128 & DecisionTree & 56.51 & 57.24 & 56.51 & 56.85 \\
    ResNet50 & 256 & RandomForest & 68.10 & 62.42 & 68.10 & 56.63 \\
    OpenAI CLIP ViT-B/32 & 256 & RandomForest & 68.02 & 61.26 & 68.02 & 56.36 \\
    OpenAI CLIP ViT-B/16 & 256 & RandomForest & 67.95 & 62.93 & 67.95 & 56.17 \\
    VGG16 & 512 & RandomForest & 67.68 & 57.22 & 67.68 & 55.90 \\
    EfficientNet-V2-L & 512 & DecisionTree & 55.19 & 56.54 & 55.19 & 55.84 \\
    LAION CLIP ViT-H/14 & 512 & RandomForest & 67.81 & 61.77 & 67.81 & 55.81 \\
    DenseNet121 & 512 & RandomForest & 67.76 & 60.47 & 67.76 & 55.74 \\
    LAION CLIP ViT-L/14 & 512 & RandomForest & 67.70 & 62.03 & 67.70 & 55.59 \\
    VGG16 & 512 & NaiveBayes & 56.61 & 56.93 & 56.61 & 55.42 \\
    OpenAI CLIP ViT-L/14 & 512 & RandomForest & 67.44 & 56.50 & 67.44 & 55.30 \\
    ResNet50 & 512 & RandomForest & 67.34 & 59.39 & 67.34 & 54.85 \\
    OpenAI CLIP ViT-B/16 & 512 & RandomForest & 67.28 & 57.27 & 67.28 & 54.69 \\
    OpenAI CLIP ViT-B/32 & 512 & RandomForest & 67.15 & 55.85 & 67.15 & 54.50 \\
    EfficientNet-V2-L & 64 & NaiveBayes & 40.00 & 62.09 & 40.00 & 44.36 \\
    EfficientNet-V2-L & 512 & NaiveBayes & 37.81 & 62.46 & 37.81 & 42.12 \\
    EfficientNet-V2-L & 128 & NaiveBayes & 32.37 & 60.36 & 32.37 & 36.46 \\
    EfficientNet-V2-L & 256 & NaiveBayes & 29.46 & 60.47 & 29.46 & 33.41 \\

\end{longtable}

%% file: long_table_2.tex
\small
\begin{small}
\begin{longtable}{llccccccccccccc}
    \caption{Summary of comparison of performances (\%) of uncertainty-aware methods using different feature extractors on HAM10000 dataset.}
    \label{tab:uncertainty_results} \\
    \hline
    \textbf{Feature Extractor} & \textbf{PCA NC} & \textbf{Method} & \textbf{Acc} & \textbf{Pre} & \textbf{Recall} & \textbf{F1} & \textbf{TU} & \textbf{IC} & \textbf{CU} & \textbf{CC} & \textbf{UAcc} & \textbf{USen} & \textbf{USpe} & \textbf{UPre} \\
    \hline
    \endfirsthead
    \hline
    \textbf{Feature Extractor} & \textbf{PCA NC} & \textbf{Method} & \textbf{Acc} & \textbf{Pre} & \textbf{Recall} & \textbf{F1} & \textbf{TU} & \textbf{IC} & \textbf{CU} & \textbf{CC} & \textbf{UAcc} & \textbf{USen} & \textbf{USpe} & \textbf{UPre} \\
    \hline
    \endhead
    \hline
    \endfoot     
    LAION CLIP ViT-H/14 & 256 & Ensemble & 85.47 & 83.90 & 85.47 & 84.54 & 231 & 60 & 394 & 1318 & 77.33 & 79.38 & 76.99 & 36.96 \\
    LAION CLIP ViT-L/14 & 256 & Ensemble & 84.62 & 83.13 & 84.62 & 83.52 & 246 & 62 & 396 & 1299 & 77.13 & 79.87 & 76.64 & 38.32 \\
    LAION CLIP ViT-L/14 & 128 & Ensemble & 84.77 & 84.25 & 84.77 & 83.85 & 247 & 58 & 400 & 1298 & 77.13 & 80.98 & 76.44 & 38.18 \\
    LAION CLIP ViT-H/14 & 128 & Ensemble & 85.02 & 83.29 & 85.02 & 83.99 & 251 & 49 & 412 & 1291 & 76.98 & 83.67 & 75.81 & 37.86 \\
    OpenAI CLIP ViT-L/14 & 256 & Ensemble & 83.72 & 81.99 & 83.72 & 82.65 & 279 & 47 & 424 & 1253 & 76.49 & 85.58 & 74.72 & 39.69 \\
    OpenAI CLIP ViT-B/16 & 256 & Ensemble & 82.88 & 81.05 & 82.88 & 81.59 & 293 & 50 & 430 & 1230 & 76.04 & 85.42 & 74.10 & 40.53 \\
    LAION CLIP ViT-H/14 & 512 & Ensemble & 85.47 & 84.11 & 85.47 & 84.55 & 235 & 56 & 424 & 1288 & 76.04 & 80.76 & 75.23 & 35.66 \\
    LAION CLIP ViT-H/14 & 64 & Ensemble & 84.07 & 83.44 & 84.07 & 83.10 & 277 & 42 & 442 & 1242 & 75.84 & 86.83 & 73.75 & 38.53 \\
    LAION CLIP ViT-L/14 & 512 & Ensemble & 84.27 & 82.60 & 84.27 & 83.17 & 262 & 53 & 433 & 1255 & 75.74 & 83.17 & 74.35 & 37.70 \\
    OpenAI CLIP ViT-B/16 & 512 & Ensemble & 82.33 & 80.63 & 82.33 & 80.95 & 294 & 60 & 431 & 1218 & 75.49 & 83.05 & 73.86 & 40.55 \\
    OpenAI CLIP ViT-L/14 & 512 & Ensemble & 83.72 & 83.45 & 83.72 & 82.74 & 276 & 50 & 447 & 1230 & 75.19 & 84.66 & 73.35 & 38.17 \\
    LAION CLIP ViT-L/14 & 64 & Ensemble & 83.77 & 83.10 & 83.77 & 82.79 & 278 & 47 & 461 & 1217 & 74.64 & 85.54 & 72.53 & 37.62 \\
    LAION CLIP ViT-L/14 & 256 & MCD & 83.82 & 82.01 & 83.82 & 82.70 & 285 & 39 & 474 & 1205 & 74.39 & 87.96 & 71.77 & 37.55 \\
    OpenAI CLIP ViT-L/14 & 128 & Ensemble & 83.87 & 83.33 & 83.87 & 82.80 & 275 & 48 & 469 & 1211 & 74.19 & 85.14 & 72.08 & 36.96 \\
    OpenAI CLIP ViT-B/32 & 256 & Ensemble & 82.28 & 80.51 & 82.28 & 80.69 & 301 & 54 & 464 & 1184 & 74.14 & 84.79 & 71.84 & 39.35 \\
    OpenAI CLIP ViT-B/16 & 128 & Ensemble & 81.78 & 79.70 & 81.78 & 80.38 & 309 & 56 & 462 & 1176 & 74.14 & 84.66 & 71.79 & 40.08 \\
    OpenAI CLIP ViT-B/32 & 128 & Ensemble & 81.78 & 79.79 & 81.78 & 80.38 & 312 & 53 & 465 & 1173 & 74.14 & 85.48 & 71.61 & 40.15 \\
    ResNet50 & 256 & Ensemble & 79.83 & 77.88 & 79.83 & 78.34 & 342 & 62 & 458 & 1141 & 74.04 & 84.65 & 71.36 & 42.75 \\
    LAION CLIP ViT-H/14 & 512 & MCD & 84.67 & 83.23 & 84.67 & 83.76 & 272 & 35 & 488 & 1208 & 73.89 & 88.60 & 71.23 & 35.79 \\
    OpenAI CLIP ViT-B/32 & 512 & Ensemble & 83.03 & 81.41 & 83.03 & 81.51 & 278 & 62 & 468 & 1195 & 73.54 & 81.76 & 71.86 & 37.27 \\
    ResNet50 & 512 & Ensemble & 79.78 & 78.06 & 79.78 & 77.99 & 347 & 58 & 478 & 1120 & 73.24 & 85.68 & 70.09 & 42.06 \\
    OpenAI CLIP ViT-B/16 & 256 & MCD & 81.08 & 78.01 & 81.08 & 79.31 & 348 & 31 & 512 & 1112 & 72.89 & 91.82 & 68.47 & 40.47 \\
    LAION CLIP ViT-L/14 & 128 & MCD & 83.13 & 81.48 & 83.13 & 82.17 & 303 & 35 & 509 & 1156 & 72.84 & 89.64 & 69.43 & 37.32 \\
    OpenAI CLIP ViT-B/32 & 64 & Ensemble & 80.03 & 77.72 & 80.03 & 78.50 & 347 & 53 & 492 & 1111 & 72.79 & 86.75 & 69.31 & 41.36 \\
    ResNet50 & 128 & Ensemble & 78.73 & 76.68 & 78.73 & 77.05 & 377 & 49 & 500 & 1077 & 72.59 & 88.50 & 68.29 & 42.99 \\
    OpenAI CLIP ViT-B/16 & 64 & Ensemble & 80.13 & 77.98 & 80.13 & 78.74 & 352 & 46 & 508 & 1097 & 72.34 & 88.44 & 68.35 & 40.93 \\
    OpenAI CLIP ViT-L/14 & 512 & MCD & 82.33 & 79.53 & 82.33 & 80.77 & 321 & 33 & 521 & 1128 & 72.34 & 90.68 & 68.41 & 38.12 \\
    VGG16 & 256 & Ensemble & 78.23 & 75.73 & 78.23 & 75.90 & 383 & 53 & 502 & 1065 & 72.29 & 87.84 & 67.96 & 43.28 \\
    LAION CLIP ViT-L/14 & 256 & EMCD & 84.17 & 82.67 & 84.17 & 82.84 & 283 & 34 & 523 & 1163 & 72.19 & 89.27 & 68.98 & 35.11 \\
    OpenAI CLIP ViT-L/14 & 64 & Ensemble & 82.83 & 80.94 & 82.83 & 81.68 & 302 & 42 & 519 & 1140 & 71.99 & 87.79 & 68.72 & 36.78 \\
    OpenAI CLIP ViT-L/14 & 256 & MCD & 82.03 & 79.10 & 82.03 & 79.95 & 325 & 35 & 533 & 1110 & 71.64 & 90.28 & 67.56 & 37.88 \\
    VGG16 & 128 & Ensemble & 77.98 & 75.31 & 77.98 & 75.77 & 391 & 50 & 520 & 1042 & 71.54 & 88.66 & 66.71 & 42.92 \\
    DenseNet121 & 256 & Ensemble & 79.38 & 76.88 & 79.38 & 77.31 & 358 & 55 & 518 & 1072 & 71.39 & 86.68 & 67.42 & 40.87 \\
    LAION CLIP ViT-L/14 & 512 & MCD & 83.62 & 81.88 & 83.62 & 82.54 & 296 & 32 & 545 & 1130 & 71.19 & 90.24 & 67.46 & 35.20 \\
    LAION CLIP ViT-H/14 & 256 & EMCD & 85.62 & 84.05 & 85.62 & 84.70 & 264 & 24 & 554 & 1161 & 71.14 & 91.67 & 67.70 & 32.27 \\
    LAION CLIP ViT-H/14 & 512 & EMCD & 85.42 & 84.07 & 85.42 & 84.47 & 261 & 31 & 547 & 1164 & 71.14 & 89.38 & 68.03 & 32.30 \\
    LAION CLIP ViT-H/14 & 128 & MCD & 84.92 & 83.22 & 84.92 & 83.92 & 276 & 26 & 554 & 1147 & 71.04 & 91.39 & 67.43 & 33.25 \\
    OpenAI CLIP ViT-B/16 & 512 & EMCD & 82.33 & 80.59 & 82.33 & 80.87 & 322 & 32 & 550 & 1099 & 70.94 & 90.96 & 66.65 & 36.93 \\
    LAION CLIP ViT-H/14 & 256 & MCD & 85.07 & 83.49 & 85.07 & 84.13 & 271 & 28 & 557 & 1147 & 70.79 & 90.64 & 67.31 & 32.73 \\
    OpenAI CLIP ViT-L/14 & 512 & EMCD & 83.42 & 82.06 & 83.42 & 82.25 & 303 & 29 & 556 & 1115 & 70.79 & 91.27 & 66.73 & 35.27 \\
    VGG16 & 64 & Ensemble & 76.98 & 73.98 & 76.98 & 74.72 & 421 & 40 & 548 & 994 & 70.64 & 91.32 & 64.46 & 43.45 \\
    OpenAI CLIP ViT-B/16 & 512 & MCD & 82.98 & 81.14 & 82.98 & 81.81 & 309 & 32 & 558 & 1104 & 70.54 & 90.62 & 66.43 & 35.64 \\
    DenseNet121 & 128 & Ensemble & 79.78 & 78.25 & 79.78 & 77.78 & 354 & 51 & 541 & 1057 & 70.44 & 87.41 & 66.15 & 39.55 \\
    LAION CLIP ViT-H/14 & 128 & EMCD & 85.12 & 83.41 & 85.12 & 84.08 & 278 & 20 & 573 & 1132 & 70.39 & 93.29 & 66.39 & 32.67 \\
    LAION CLIP ViT-L/14 & 512 & EMCD & 84.27 & 82.64 & 84.27 & 83.15 & 286 & 29 & 566 & 1122 & 70.29 & 90.79 & 66.47 & 33.57 \\
    VGG16 & 512 & Ensemble & 78.73 & 77.48 & 78.73 & 75.75 & 370 & 56 & 539 & 1038 & 70.29 & 86.85 & 65.82 & 40.70 \\
    ResNet50 & 256 & MCD & 78.68 & 76.56 & 78.68 & 77.27 & 391 & 36 & 562 & 1014 & 70.14 & 91.57 & 64.34 & 41.03 \\
    DenseNet121 & 64 & Ensemble & 78.28 & 76.78 & 78.28 & 75.93 & 387 & 48 & 551 & 1017 & 70.09 & 88.97 & 64.86 & 41.26 \\
    LAION CLIP ViT-L/14 & 128 & EMCD & 84.72 & 84.18 & 84.72 & 83.78 & 274 & 32 & 568 & 1129 & 70.04 & 89.54 & 66.53 & 32.54 \\
    DenseNet121 & 512 & Ensemble & 80.08 & 77.57 & 80.08 & 77.95 & 349 & 50 & 551 & 1053 & 70.00 & 87.47 & 65.65 & 38.78 \\
    ResNet50 & 64 & Ensemble & 77.98 & 75.32 & 77.98 & 76.07 & 391 & 50 & 552 & 1010 & 69.95 & 88.66 & 64.66 & 41.46 \\
    OpenAI CLIP ViT-L/14 & 256 & EMCD & 83.87 & 82.19 & 83.87 & 82.81 & 291 & 32 & 574 & 1106 & 69.75 & 90.09 & 65.83 & 33.64 \\
    OpenAI CLIP ViT-B/32 & 256 & MCD & 81.58 & 79.61 & 81.58 & 79.97 & 332 & 37 & 572 & 1062 & 69.60 & 89.97 & 64.99 & 36.73 \\
    OpenAI CLIP ViT-B/16 & 256 & EMCD & 82.73 & 80.92 & 82.73 & 81.40 & 315 & 31 & 579 & 1078 & 69.55 & 91.04 & 65.06 & 35.23 \\
    ResNet50 & 512 & MCD & 78.58 & 75.50 & 78.58 & 76.88 & 392 & 37 & 580 & 994 & 69.20 & 91.38 & 63.15 & 40.33 \\
    ResNet50 & 512 & EMCD & 79.63 & 77.85 & 79.63 & 77.78 & 377 & 31 & 588 & 1007 & 69.10 & 92.40 & 63.13 & 39.07 \\
    LAION CLIP ViT-H/14 & 64 & MCD & 83.28 & 82.24 & 83.28 & 82.34 & 306 & 29 & 593 & 1075 & 68.95 & 91.34 & 64.45 & 34.04 \\
    ResNet50 & 256 & EMCD & 79.73 & 77.76 & 79.73 & 78.16 & 371 & 35 & 588 & 1009 & 68.90 & 91.38 & 63.18 & 38.69 \\
    OpenAI CLIP ViT-B/32 & 512 & MCD & 81.58 & 78.50 & 81.58 & 79.41 & 335 & 34 & 589 & 1045 & 68.90 & 90.79 & 63.95 & 36.26 \\
    OpenAI CLIP ViT-B/32 & 512 & EMCD & 82.78 & 81.09 & 82.78 & 81.03 & 314 & 31 & 594 & 1064 & 68.80 & 91.01 & 64.17 & 34.58 \\
    ResNet50 & 128 & MCD & 77.78 & 75.59 & 77.78 & 76.09 & 412 & 33 & 592 & 966 & 68.80 & 92.58 & 62.00 & 41.04 \\
    OpenAI CLIP ViT-B/32 & 256 & EMCD & 82.28 & 80.50 & 82.28 & 80.62 & 323 & 32 & 594 & 1054 & 68.75 & 90.99 & 63.96 & 35.22 \\
    LAION CLIP ViT-L/14 & 64 & MCD & 83.52 & 82.81 & 83.52 & 82.48 & 299 & 31 & 600 & 1073 & 68.50 & 90.61 & 64.14 & 33.26 \\
    OpenAI CLIP ViT-B/32 & 128 & MCD & 81.83 & 79.79 & 81.83 & 80.49 & 335 & 29 & 604 & 1035 & 68.40 & 92.03 & 63.15 & 35.68 \\
    OpenAI CLIP ViT-L/14 & 128 & MCD & 82.88 & 82.02 & 82.88 & 81.76 & 310 & 33 & 603 & 1057 & 68.25 & 90.38 & 63.67 & 33.95 \\
    LAION CLIP ViT-H/14 & 64 & EMCD & 83.62 & 83.02 & 83.62 & 82.59 & 302 & 26 & 615 & 1060 & 68.00 & 92.07 & 63.28 & 32.93 \\
    OpenAI CLIP ViT-B/16 & 128 & EMCD & 81.63 & 79.51 & 81.63 & 80.16 & 335 & 33 & 611 & 1024 & 67.85 & 91.03 & 62.63 & 35.41 \\
    OpenAI CLIP ViT-B/16 & 128 & MCD & 80.88 & 78.96 & 80.88 & 79.58 & 349 & 34 & 610 & 1010 & 67.85 & 91.12 & 62.35 & 36.39 \\
    OpenAI CLIP ViT-L/14 & 128 & EMCD & 83.72 & 81.96 & 83.72 & 82.56 & 299 & 27 & 617 & 1060 & 67.85 & 91.72 & 63.21 & 32.64 \\
    LAION CLIP ViT-L/14 & 64 & EMCD & 83.82 & 83.16 & 83.82 & 82.78 & 296 & 28 & 619 & 1060 & 67.70 & 91.36 & 63.13 & 32.35 \\
    OpenAI CLIP ViT-B/32 & 128 & EMCD & 81.98 & 79.91 & 81.98 & 80.50 & 331 & 30 & 625 & 1017 & 67.30 & 91.69 & 61.94 & 34.62 \\
    VGG16 & 256 & MCD & 78.08 & 75.38 & 78.08 & 75.74 & 402 & 37 & 619 & 945 & 67.25 & 91.57 & 60.42 & 39.37 \\
    VGG16 & 128 & MCD & 77.08 & 74.22 & 77.08 & 74.97 & 418 & 41 & 618 & 926 & 67.10 & 91.07 & 59.97 & 40.35 \\
    ResNet50 & 128 & EMCD & 78.73 & 76.66 & 78.73 & 76.92 & 399 & 27 & 633 & 944 & 67.05 & 93.66 & 59.86 & 38.66 \\
    OpenAI CLIP ViT-B/32 & 64 & EMCD & 80.28 & 78.00 & 80.28 & 78.74 & 373 & 22 & 639 & 969 & 67.00 & 94.43 & 60.26 & 36.86 \\
    OpenAI CLIP ViT-B/32 & 64 & MCD & 79.53 & 77.16 & 79.53 & 77.92 & 376 & 34 & 631 & 962 & 66.80 & 91.71 & 60.39 & 37.34 \\
    VGG16 & 512 & MCD & 77.08 & 71.44 & 77.08 & 73.74 & 420 & 39 & 629 & 915 & 66.65 & 91.50 & 59.26 & 40.04 \\
    DenseNet121 & 128 & MCD & 78.83 & 76.02 & 78.83 & 76.78 & 393 & 31 & 639 & 940 & 66.55 & 92.69 & 59.53 & 38.08 \\
    VGG16 & 64 & MCD & 76.19 & 74.31 & 76.19 & 74.23 & 455 & 22 & 648 & 878 & 66.55 & 95.39 & 57.54 & 41.25 \\
    EfficientNet-V2-L & 512 & Ensemble & 71.24 & 64.14 & 71.24 & 65.47 & 531 & 45 & 627 & 800 & 66.45 & 92.19 & 56.06 & 45.85 \\
    OpenAI CLIP ViT-L/14 & 64 & MCD & 82.08 & 80.20 & 82.08 & 80.92 & 331 & 28 & 646 & 998 & 66.35 & 92.20 & 60.71 & 33.88 \\
    OpenAI CLIP ViT-B/16 & 64 & MCD & 80.48 & 78.38 & 80.48 & 79.10 & 357 & 34 & 651 & 961 & 65.80 & 91.30 & 59.62 & 35.42 \\
    DenseNet121 & 64 & MCD & 76.29 & 74.23 & 76.29 & 74.00 & 456 & 19 & 668 & 860 & 65.70 & 96.00 & 56.28 & 40.57 \\
    OpenAI CLIP ViT-L/14 & 64 & EMCD & 82.78 & 80.88 & 82.78 & 81.61 & 320 & 25 & 663 & 995 & 65.65 & 92.75 & 60.01 & 32.55 \\
    DenseNet121 & 256 & MCD & 78.48 & 76.08 & 78.48 & 76.72 & 402 & 29 & 661 & 911 & 65.55 & 93.27 & 57.95 & 37.82 \\
    OpenAI CLIP ViT-B/16 & 64 & EMCD & 80.38 & 78.27 & 80.38 & 78.98 & 365 & 28 & 662 & 948 & 65.55 & 92.88 & 58.88 & 35.54 \\
    VGG16 & 128 & EMCD & 77.88 & 75.06 & 77.88 & 75.37 & 414 & 29 & 663 & 897 & 65.45 & 93.45 & 57.50 & 38.44 \\
    DenseNet121 & 512 & MCD & 78.28 & 76.02 & 78.28 & 76.24 & 399 & 36 & 663 & 905 & 65.10 & 91.72 & 57.72 & 37.57 \\
    VGG16 & 256 & EMCD & 78.08 & 75.33 & 78.08 & 75.53 & 410 & 29 & 673 & 891 & 64.95 & 93.39 & 56.97 & 37.86 \\
    ResNet50 & 64 & EMCD & 77.88 & 75.22 & 77.88 & 75.90 & 413 & 30 & 672 & 888 & 64.95 & 93.23 & 56.92 & 38.06 \\
    ResNet50 & 64 & MCD & 78.13 & 75.61 & 78.13 & 76.24 & 404 & 34 & 671 & 894 & 64.80 & 92.24 & 57.12 & 37.58 \\
    VGG16 & 512 & EMCD & 78.38 & 77.14 & 78.38 & 75.21 & 399 & 34 & 671 & 899 & 64.80 & 92.15 & 57.26 & 37.29 \\
    EfficientNet-V2-L & 256 & Ensemble & 71.94 & 66.17 & 71.94 & 67.49 & 516 & 46 & 665 & 776 & 64.50 & 91.81 & 53.85 & 43.69 \\
    VGG16 & 64 & EMCD & 76.93 & 73.79 & 76.93 & 74.41 & 442 & 20 & 701 & 840 & 64.00 & 95.67 & 54.51 & 38.67 \\
    DenseNet121 & 64 & EMCD & 78.13 & 75.47 & 78.13 & 75.70 & 424 & 14 & 713 & 852 & 63.70 & 96.80 & 54.44 & 37.29 \\
    DenseNet121 & 128 & EMCD & 79.38 & 76.70 & 79.38 & 77.21 & 392 & 21 & 715 & 875 & 63.26 & 94.92 & 55.03 & 35.41 \\
    DenseNet121 & 512 & EMCD & 79.73 & 77.04 & 79.73 & 77.41 & 380 & 26 & 714 & 883 & 63.06 & 93.60 & 55.29 & 34.73 \\
    DenseNet121 & 256 & EMCD & 79.63 & 77.11 & 79.63 & 77.48 & 385 & 23 & 719 & 876 & 62.96 & 94.36 & 54.92 & 34.87 \\
    EfficientNet-V2-L & 128 & Ensemble & 72.24 & 66.50 & 72.24 & 67.30 & 516 & 40 & 717 & 730 & 62.21 & 92.81 & 50.45 & 41.85 \\
    EfficientNet-V2-L & 512 & EMCD & 71.09 & 64.04 & 71.09 & 65.15 & 555 & 24 & 745 & 679 & 61.61 & 95.85 & 47.68 & 42.69 \\
    EfficientNet-V2-L & 512 & MCD & 70.84 & 60.62 & 70.84 & 64.52 & 562 & 22 & 751 & 668 & 61.41 & 96.23 & 47.08 & 42.80 \\
    EfficientNet-V2-L & 64 & Ensemble & 71.54 & 65.00 & 71.54 & 65.79 & 542 & 28 & 755 & 678 & 60.91 & 95.09 & 47.31 & 41.79 \\
    EfficientNet-V2-L & 256 & MCD & 72.14 & 66.67 & 72.14 & 68.16 & 525 & 33 & 774 & 671 & 59.71 & 94.09 & 46.44 & 40.42 \\
    EfficientNet-V2-L & 128 & MCD & 72.14 & 67.25 & 72.14 & 67.84 & 535 & 23 & 799 & 646 & 58.96 & 95.88 & 44.71 & 40.10 \\
    EfficientNet-V2-L & 256 & EMCD & 72.04 & 66.15 & 72.04 & 67.44 & 532 & 28 & 799 & 644 & 58.71 & 95.00 & 44.63 & 39.97 \\
    EfficientNet-V2-L & 128 & EMCD & 72.24 & 66.48 & 72.24 & 67.13 & 537 & 19 & 842 & 605 & 57.01 & 96.58 & 41.81 & 38.94 \\
    EfficientNet-V2-L & 64 & MCD & 70.99 & 63.45 & 70.99 & 64.75 & 558 & 23 & 839 & 583 & 56.96 & 96.04 & 41.00 & 39.94 \\
    EfficientNet-V2-L & 64 & EMCD & 71.44 & 64.79 & 71.44 & 65.56 & 556 & 16 & 878 & 553 & 55.37 & 97.20 & 38.64 & 38.77 \\
    \hline    
\end{longtable}
\end{small}